\documentclass[twoside,11pt]{article}

\usepackage{jmlr2e}
\usepackage{amsmath}
\usepackage{booktabs}
\usepackage{enumerate}
\usepackage{hyperref}
\newcommand\numberthis{\addtocounter{equation}{1}\tag{\theequation}}

\newcommand{\fracpartial}[2]{\frac{\partial #1}{\partial  #2}}
\newcommand{\norm}[1]{\left\lVert#1\right\rVert}

\jmlrheading{VV}{20YY}{PP}{06/2019}{MM/20YY}{marschall00a}{Owen Marschall, Kyunghyun Cho, and Cristina Savin}

\ShortHeadings{A Unified Framework of Online Learning Algorithms}{Marschall, Cho, and Savin}
\firstpageno{1}

\begin{document}

\title{A Unified Framework of Online Learning Algorithms for Training Recurrent Neural Networks}

\author{\name Owen Marschall \email oem214@nyu.edu \\
       \addr Center for Neural Science\\
       New York University\\
       New York, NY 10006, USA
       \AND
       \name Kyunghyun Cho \email kyunghyun.cho@nyu.edu \\
       \addr 
       New York University\\
       Facebook AI Research \\
       CIFAR Azrieli Global Scholar
       \AND
       \name Cristina Savin \email csavin@nyu.edu \\
       \addr Center for Neural Science\\
       Center for Data Science\\
       New York University\\}

\editor{ }

\maketitle

\begin{abstract}
We present a framework for compactly summarizing many recent results in efficient and/or biologically plausible online training of recurrent neural networks (RNN). The framework organizes algorithms according to several criteria: (a) past vs.\ future facing, (b) tensor structure, (c) stochastic vs.\ deterministic, and (d) closed form vs.\ numerical. These axes reveal latent conceptual connections among several recent advances in online learning. Furthermore, we provide novel mathematical intuitions for their degree of success. Testing various algorithms on two synthetic tasks shows that performances cluster according to our criteria. Although a similar clustering is also observed for gradient alignment, alignment with exact methods does not alone explain ultimate performance, especially for stochastic algorithms. This suggests the need for better comparison metrics.
\end{abstract}


\section{Introduction}

Training recurrent neural networks (RNN) to learn sequence data is traditionally done with stochastic gradient descent (SGD), using the backpropagation through time algorithm (BPTT, \citealp{werbos1990backpropagation}) to calculate the gradient. This requires ``unrolling'' the network over some range of time steps $T$ and performing backpropagation as though the network were feedforward under the constraint of sharing parameters across time steps (``layers''). BPTT's success in a wide range of applications \citep{mikolov2010recurrent,graves2013generating,bahdanau2016end,bahdanau2014neural,cho2015describing,graves2016hybrid} has made it the industry standard; however, there exist alternative {\bf online} algorithms for training RNNs. These compute gradients in real time as the network runs forward, without explicitly referencing past activity or averaging over batches of data. There are two reasons for considering online alternatives to BPTT. One is practical: computational costs do not scale with $T$. The other is conceptual: human brains are able to learn long-term dependencies without explicitly memorizing all past brain states, and understanding online learning is a key step in the larger project of understanding human learning.

The classic online learning algorithm is real-time recurrent learning (RTRL, \citealp{williams1989learning}), which is equivalent to BPTT in the limit of a small learning rate (\citealp{murray2019local}). RTRL recursively updates the total derivative of the hidden state with respect to the parameters, eliminating the need to reference past activity but introducing an order \mbox{$n \text{ (hidden units) } \times n^2 \text{ (parameters) } = n^3$} memory requirement. In practice, this is often more computationally demanding than BPTT (order $nT$), hence not frequently used in applications. Nor is RTRL at face value a good model of biological learning, for the same reason: no known biological mechanism exists to store---let alone manipulate---a float for each synapse-neuron pair. Thus RTRL and online learning more broadly have remained relatively obscure footnotes to both the deep learning revolution itself and its impact on computational neuroscience.

Recent advances in recurrent network architectures have brought the issue of online learning back into the spotlight. While vanishing gradients used to significantly limit the extent of the temporal dependencies that an RNN could learn, new architectures like LSTMs \citep{hochreiter1997long} and GRUs \citep{cho2014learning} have dramatically expanded this learnable time horizon. Unfortunately, taking advantage of this capacity requires an equally dramatic expansion in computational resources, if using BPTT. This has led to an explosion of novel online learning algorithms \citep{tallec2017unbiased, mujika2018approximating, roth2018kernel, murray2019local, jaderberg2017decoupled} which aim to improve on the efficiency of RTRL, in many cases using update rules that might be implemented by a biological circuit.

The sheer number and variety of these approaches pose challenges for both theory and practice. It is not always completely clear what makes various algorithms different from one another, how they are conceptually related, or even why they might work in the first place. There is a pressing need in the field for a cohesive framework for describing and comparing online methods. Here we aim to provide a thorough overview of modern online algorithms for training RNNs, in a way that provides a clearer understanding of the mathematical structure underlying all these different approaches. Our framework organizes the existing literature along several axes that encode meaningful conceptual distinctions:
\begin{enumerate}[a)]
\itemsep0em
    \item {\bf Past facing} vs.\ {\bf future facing}
    \item The {\bf tensor structure} of the algorithm
    \item {\bf Stochastic} vs.\ {\bf deterministic} update
    \item {\bf Closed form} vs.\ {\bf numerical} solution for update
\end{enumerate}
These axes will be explained in detail later, but briefly: the past vs.\ future axis is a root distinction that divides algorithms by the type of gradient they calculate, while the other three describe their representations and update principles. Table~\ref{algtable} contains (to our knowledge) all recently published online learning algorithms for RNNs, categorized according to these criteria. We can already see that many combinations of these characteristics manifest in the literature, suggesting that new algorithms could be developed by mixing and matching properties. (We provide a concrete example of this in \S\ref{ikf-rtrl}.)
\begin{table}[t]
\begin{tabular}{l|l|l|l|l|l|l}
\textbf{Algorithm} & \textbf{Facing} & \textbf{Tensor}     & \multicolumn{2}{c|}{\textbf{Update}} & \textbf{Memory} & \textbf{Time} \\ 
\toprule
RTRL               & Past            & $M_{kij}$                     & Deterministic        & Closed-form              & $n^3$           & $n^4$                \\ 
UORO               & Past            & $A_k B_{ij}$                  & Stochastic           & Closed-form              & $n^2$           & $n^2$                \\ 
KF-RTRL            & Past            & $A_j B_{ki}$                  & Stochastic           & Closed-form              & $n^2$           & $n^3$                \\ 
R-KF-RTRL            & Past            & $A_i B_{jk}$                  & Stochastic           & Closed-form              & $n^2$           & $n^3$                \\ 
$r$-OK             & Past            & $\sum_{l=1}^r A_{lj} B_{lki}$ & Stochastic           & Numerical                & $rn^2$          & $rn^3$               \\ 
KeRNL              & Past            & $A_{ki} B_{ij}$               & Deterministic        & Numerical                     & $n^2$           & $n^2$                \\ 
RFLO               & Past            & $\delta_{ki} B_{ij}$          & Deterministic        & Closed-form              & $n^2$           & $n^2$                \\ 
\midrule
 
E-BPTT            & --          & --                           & Deterministic        & Closed-form              & $nT$            & $n^2$               \\ 
F-BPTT            & Future          & --                           & Deterministic        & Closed-form              & $nT$            & $n^2T$               \\
DNI                & Future          & $A_{li}$                      & Deterministic        & Numerical                & $n^2$           & $n^2$                \\ 
\end{tabular}
\caption{A list of learning algorithms reviewed here, together with their main properties. The indices $k$, $i$ and $j$ reference different dimensions of the ``influence tensor" of RTRL (\S\ref{rtrl_approx}); $l$ references components of the feedback vector ${\bf \tilde{a}}$ in DNI (\S\ref{dni}).}
\label{algtable}
\end{table}

Here we describe each algorithm in unified notation that makes clear their classification by these criteria. In the process, we generate novel intuitions about why different approximations can be successful and discuss some of the finer points of their biological plausibility. Finally, we simulate each algorithm on a common set of synthetic tasks with vanilla RNN architecture for simplicity. We compare performance and analyze gradient alignments to see to what extent their empirical similarity is predicted by their similarity according to our framework. Algorithm performance roughly clusters according to criteria (a)-(d) across tasks, lending credence to our approach. Curiously, gradient alignment with exact methods (RTRL and BPTT) does not predict performance, despite its ubiquity as a tool for analyzing approximate learning algorithms.

\section{Past- and future-facing perspectives of online learning}
\label{pf_ff_online_learning}

Before we dive into the details of these algorithms, we first articulate what we mean by past- and future-facing, related to the ``reverse/forward accumulation" distinction concurrently described by \cite{cooijmans2019variance}. Consider a recurrent neural network that contains, at each time step $t$, a state ${\bf a}^{(t)} \in \mathbb{R}^n$. This state is updated via a function \mbox{$F_{{\bf w}}: \mathbb{R}^{m} \rightarrow \mathbb{R}^{n}$}, which is parameterized by a flattened vector of parameters ${\bf w} \in \mathbb{R}^P$. Here $m = n + n_{in} + 1$ counts the total number of input dimensions, including the recurrent inputs ${\bf a}^{(t-1)} \in \mathbb{R}^n$, task inputs ${\bf x}^{(t)} \in \mathbb{R}^{n_{\text{in}}}$, and an additional input clamped to $1$ (to represent bias). For some initial state ${\bf a}^{(0)}$, $F_{{\bf w}}$ defines the network dynamics by
\begin{equation*}
{\bf a}^{(t)} = F_{{\bf w}}({\bf a}^{(t-1)}, {\bf x}^{(t)}).
\label{abstract_rnn_eq}
\end{equation*}
At each time step an output ${\bf y}^{(t)} \in \mathbb{R}^{n_{\text{out}}}$ is computed by another function $F_{{\bf w}_{\text{o}}}^{\text{out}}: \mathbb{R}^{n} \rightarrow \mathbb{R}^{n_{out}}$, parameterized by ${\bf w}_{\text{o}} \in \mathbb{R}^{P_{\text{o}}}$. We will typically choose an affine-softmax readout for $F^{\text{out}}_{{\bf w}_{\text{o}}}$, with output weights/bias ${\bf W}^{\text{out}} \in \mathbb{R}^{n_{\text{out}} \times (n+1)}$. A loss function $L({\bf y}^{(t)}, {\bf y}^{*(t)})$ calculates an instantaneous loss $L^{(t)}$, quantifying to what degree the predicted output ${\bf y}^{(t)}$ matches the target output ${\bf y}^{*(t)}$. 

The goal is to train the network by gradient descent (or other gradient-based optimizers such as ADAM from \citealp{kingma2014adam}) on the total loss $\mathcal{L} = \sum L^{(t)}$ w.r.t. the parameters ${\bf w}$ and ${\bf w}_{\text{o}}$. It is natural to learn ${\bf w}_{\text{o}}$ online, because only information at present time $t$ is required to calculate the gradient $\partial L^{(t)} / \partial {\bf w}_{\text{o}}$. So the heart of the problem is to calculate $\partial \mathcal{L} / \partial {\bf w}$.

\begin{figure}[t]
\includegraphics[width=\textwidth]{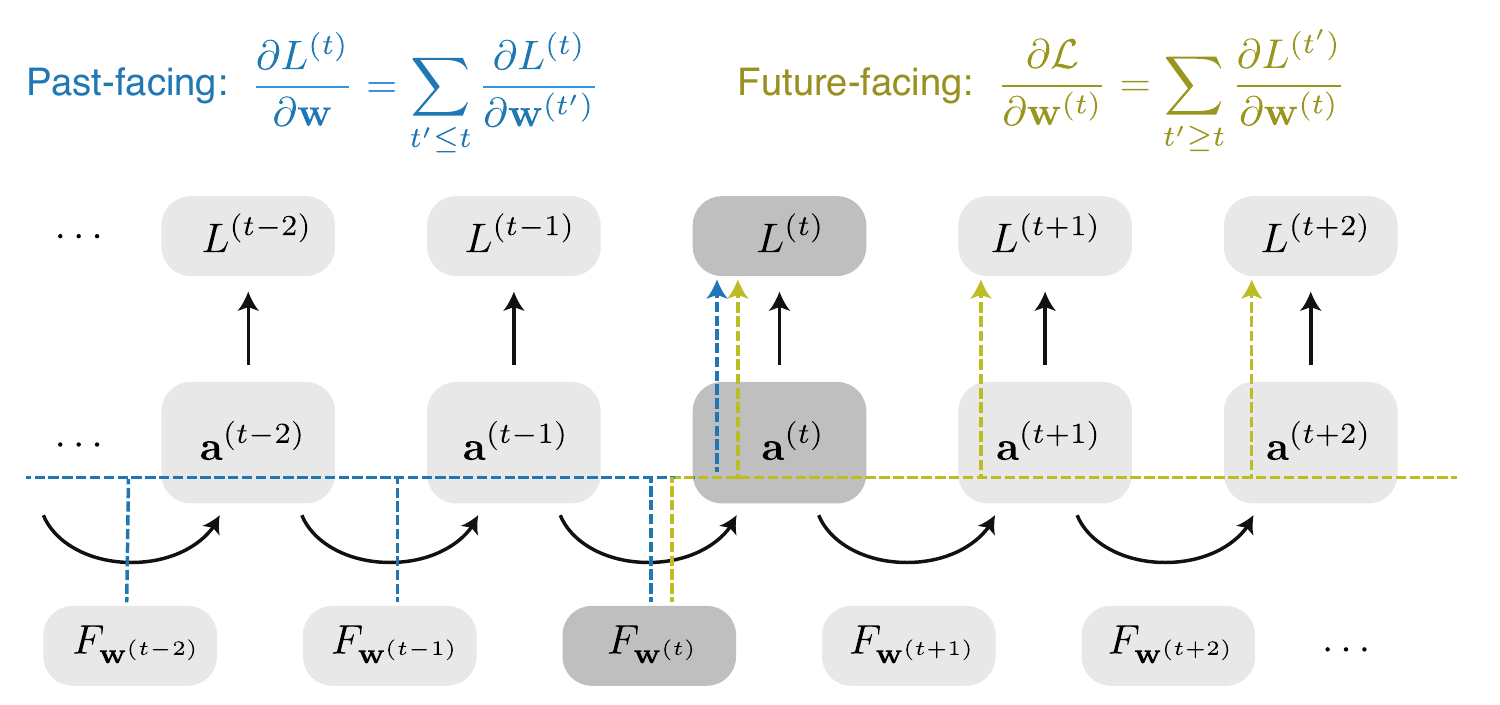}
\caption{Cartoon depicting the past- and future-facing perspectives of online learning, for an RNN unrolled over time. Each ${\bf a}$ represents the RNN hidden state value, while $F_{\bf w}$ denotes applications of the recurrent update; the instantaneous losses $L$ implicitly depend on the hidden state through $L^{(t)} = L\left(F^{\text{out}}_{{\bf w}_{\text{o}}}({\bf a}^{(t)}), {\bf y}^{*(t)}\right)$. The blue (yellow) arrows show the paths of influence accounted for by the past-facing (future-facing) gradient described in the corresponding equation.}
\label{fig:past_future}
\end{figure}

The parameter ${\bf w}$ is applied via $F_{{\bf w}}$ at every time step, and we denote a particular application of ${\bf w}$ at time $s$ as ${\bf w}^{(s)}$. Of course, a recurrent system is constrained to share parameters across time steps, so a perturbation $\delta {\bf w}$ is effectively a perturbation across all applications $\delta {\bf w}^{(s)}$, i.e., $\partial {\bf w}^{(s)} / \partial {\bf w} = {\bf I}_P$. In principle, each application of the parameters affects all future losses $L^{(t)}$, $t \geq s$. The core of any recurrent learning algorithm is to estimate the influence $\partial L^{(t)} / \partial {\bf w}^{(s)}$ of one parameter application ${\bf w}^{(s)}$ on one loss $L^{(t)}$, since these individual terms are necessary and sufficient to define the global gradient
\begin{equation}
\frac{\partial \mathcal{L}}{\partial {\bf w}} = \sum_{t} \frac{\partial L^{(t)}}{\partial {\bf w}} = \sum_t \sum_{s \leq t} \frac{\partial L^{(t)}}{\partial {\bf w}^{(s)}}\frac{\partial {\bf w}^{(s)}}{\partial {\bf w}} =  \sum_t \sum_{s \leq t} \frac{\partial L^{(t)}}{\partial {\bf w}^{(s)}}.
\label{loss_sum}
\end{equation}
This raises the question of how to sum these components to produce individual gradients to pass to the optimizer. In truncated BPTT, one unrolls the graph over some range of time steps and sums $\partial L^{(t)} / \partial {\bf w}^{(s)}$ for all $t, s$ in that range with $t \geq s$ (see \S\ref{e-bptt}). This does not qualify as an ``online" learning rule, because it requires two independent time indices---at most one can represent ``real time'' leaving the other to represent the future or the past. If we can account for one of the summations via dynamic updates, then the algorithm is {\bf online} or {\bf temporally local}, i.e. not requiring explicit reference to the past or future. As depicted in Fig.~\ref{fig:past_future}, there are two possibilities. If $t$ from Eq.~\eqref{loss_sum} corresponds to real time, then the gradient passed to the optimizer is
\begin{equation}
\nabla_{{\bf w}} \mathcal{L}(t) = \sum_{s=0}^t \frac{\partial L^{(t)}}{\partial {\bf w}^{(s)}} = \frac{\partial L^{(t)}}{\partial {\bf w}}.
\label{past_facing_sum}
\end{equation}
In this case, we say learning is {\bf past facing}, because the gradient is a sum of the influences of \emph{past} applications of ${\bf w}$ on the current loss. On the other hand, if $s$ from Eq.~\eqref{loss_sum} represents real time, then the gradient passed to the optimizer is
\begin{equation}
\nabla_{{\bf w}} \mathcal{L}(s) = \sum_{t=s}^{\infty} \frac{\partial L^{(t)}}{\partial {\bf w}^{(s)}} = \frac{\partial \mathcal{L}}{\partial {\bf w}^{(s)}}.
\label{future_facing_sum}
\end{equation}
Here we say learning is {\bf future facing}, because the gradient is a sum of influences by the current application of ${\bf w}$ on \emph{future} losses.

\subsection{Past-facing online learning algorithms}
Here we derive a fundamental relation leveraged by past-facing (PF) online algorithms. Let $t$ index real time, and define the {\bf influence matrix} ${\bf M}^{(t)} \in \mathbb{R}^{n \times P}$, where $n$ and $P$ are respectively the number of hidden units and the number of parameters defining $F_{{\bf w}}$. ${\bf M}^{(t)}$ tracks the derivatives of the current state ${\bf a}^{(t)}$ with respect to each parameter $w_p$:
\begin{equation}
M^{(t)}_{kp} = \frac{\partial a^{(t)}_k}{\partial w_{p}}.
\label{def_influence}
\end{equation}
Let's rewrite Eq.~\eqref{def_influence} with matrix notation and unpack it by one time step:
\begin{align*}
{\bf M}^{(t)} &= \frac{\partial {\bf a}^{(t)}}{\partial {\bf w}} = \sum_{s \leq t} \frac{\partial {\bf a}^{(t)}}{\partial {\bf w}^{(s)}} = \sum_{s \leq t-1} \frac{\partial {\bf a}^{(t)}}{\partial {\bf w}^{(s)}} + \frac{\partial {\bf a}^{(t)}}{\partial {\bf w}^{(t)}}\\
&= \sum_{s \leq t-1} \frac{\partial {\bf a}^{(t)}}{\partial {\bf a}^{(t-1)}}\frac{\partial {\bf a}^{(t-1)}}{\partial {\bf w}^{(s)}} + \frac{\partial {\bf a}^{(t)}}{\partial {\bf w}^{(t)}}\\
&= \frac{\partial {\bf a}^{(t)}}{\partial {\bf a}^{(t-1)}}\frac{\partial {\bf a}^{(t-1)}}{\partial {\bf w}} + \frac{\partial {\bf a}^{(t)}}{\partial {\bf w}^{(t)}}\\
&\equiv {\bf J}^{(t)}{\bf M}^{(t-1)} + {\bf \overline{M}}^{(t)}. \numberthis
\label{pf_relation}
\end{align*}
A simple recursive formula emerges, wherein the influence matrix is updated by multiplying its current value by the Jacobian ${\bf J}^{(t)} = \partial {\bf a}^{(t)} / \partial {\bf a}^{(t-1)} \in \mathbb{R}^{n \times n}$ of the network and then adding the {\bf immediate influence} ${\bf \overline{M}}^{(t)} = \partial {\bf a}^{(t)} / \partial {\bf w}^{(t)} \in \mathbb{R}^{n \times P}$. To compute the gradient that ultimately gets passed to the optimizer, we simply use the chain rule over the current hidden state ${\bf a}^{(t)}$:
\begin{equation}
\frac{\partial L^{(t)}}{\partial {\bf w}} = \frac{\partial L^{(t)}}{\partial {\bf a}^{(t)}}\frac{\partial {\bf a}^{(t)}}{\partial {\bf w}} \equiv {\bf \overline{c}}^{(t)}{\bf M}^{(t)},
\label{pf_grad}
\end{equation}
where the {\bf immediate credit assignment vector} ${\bf \overline{c}}^{(t)} \in \mathbb{R}^n$ is defined to be $\partial L^{(t)} / \partial {\bf a}^{(t)}$ and is calculated by backpropagating the error $\boldsymbol{\delta}^{(t)}$ through the derivative of the output function $F^{\text{out}}_{{\bf w}_{\text{o}}}$ (or approximated by Feedback Alignment, see \citealp{lillicrap2016random}). In the end, we compute a derivative in Eq.~\eqref{pf_grad} that is implicitly a sum over the many terms of Eq.~\eqref{past_facing_sum}, using formulae that depend explicitly only on times $t$ and $t-1$. For this reason, such a learning algorithm is {\bf online}, and it is {\bf past facing} because the gradient computation is of the form in Eq.~\eqref{past_facing_sum}.
\subsection{Future-facing online learning algorithms}
Here we show a symmetric relation for future-facing (FF) online algorithms. The {\bf credit assignment vector} ${\bf c}^{(t)} \in \mathbb{R}^n$ is a row vector defined as the gradient of the loss $\mathcal{L}$ with respect to the hidden state ${\bf a}^{(t)}$. It plays a role analogous to ${\bf M}^{(t)}$ and has a recursive update similar to Eq.~\eqref{pf_relation}:
\begin{align*}
{\bf c}^{(t)} &= \frac{\partial \mathcal{L}}{\partial {\bf a}^{(t)}} = \sum_{s \geq t}\frac{\partial L^{(s)}}{\partial {\bf a}^{(t)}} = \frac{\partial L^{(t)}}{\partial {\bf a}^{(t)}} + \sum_{s \geq t+1}\frac{\partial L^{(s)}}{\partial {\bf a}^{(t)}}\\
&= \frac{\partial L^{(t)}}{\partial {\bf a}^{(t)}} + \sum_{s \geq t+1}\frac{\partial L^{(s)}}{\partial {\bf a}^{(t+1)}}\frac{\partial {\bf a}^{(t+1)}}{\partial {\bf a}^{(t)}}\\
&= \frac{\partial L^{(t)}}{\partial {\bf a}^{(t)}} + \frac{\partial \mathcal{L}}{\partial {\bf a}^{(t+1)}}\frac{\partial {\bf a}^{(t+1)}}{\partial {\bf a}^{(t)}}\\
&= {\bf \overline{c}}^{(t)} + {\bf c}^{(t+1)}{\bf J}^{(t+1)}. \numberthis
\label{ff_relation}
\end{align*}
As in the PF case, the gradient is ultimately calculated using the chain rule over ${\bf a}^{(t)}$:
\begin{equation}
\frac{\partial \mathcal{L}}{\partial {\bf w}^{(t)}} = \frac{\partial \mathcal{L}}{\partial {\bf a}^{(t)}}\frac{\partial {\bf a}^{(t)}}{\partial {\bf w}^{(t)}} \equiv {\bf c}^{(t)}{\bf \overline{M}}^{(t)}.
\label{ff_gradient}
\end{equation}
The recursive relations for PF and FF algorithms are of identical form given the following changes: (1) swap the roles of $\mathcal{L}$ and ${\bf w}$, (2) swap the roles of $t-1$ and $t+1$, and (3) flip the direction of all derivatives. This clarifies the fundamental trade-off between the PF and FF approaches to online learning. On the one hand, memory requirements favor FF because $\mathcal{L}$ is a scalar while ${\bf w}$ is a matrix. On the other, only PF can truly be run online, because the time direction of the update in FF is opposite the forward pass. Thus, efficient PF algorithms must \emph{compress} ${\bf M}^{(t)}$, while efficient FF algorithms must \emph{predict} ${\bf c}^{(t+1)}$.
\section{Past-facing algorithms}
\subsection{Real-Time Recurrent Learning}
The Real-Time Recurrent Learning (RTRL, \citealp{williams1989learning}) algorithm directly applies Eqs. \eqref{pf_relation} and \eqref{pf_grad} as written. We call the application of Eq.~\eqref{pf_relation} the ``update" to the learning algorithm, which is {\bf deterministic} and in {\bf closed form}. Implementing Eq.~\eqref{pf_relation} requires storing $nP \approx \mathcal{O}(n^3)$ floats in ${\bf M}^{(t)}$ and performing $\mathcal{O}(n^4)$ multiplications in ${\bf J}^{(t)}{\bf M}^{(t)}$, which is neither especially efficient nor biologically plausible. However, several efficient (and in some cases, biologically plausible) online learning algorithms have recently been developed, including Unbiased Online Recurrent Optimization (UORO; \citealp{tallec2017unbiased}; \S\ref{uoro}), Kronecker-Factored RTRL (KF-RTRL; \citealp{mujika2018approximating}; \S\ref{kf-rtrl}), Kernel RNN Learning (KeRNL; \citealp{roth2018kernel}; \S\ref{kernl}), and Random-Feedback Online Learning (RFLO; \citealp{murray2019local}; \S\ref{rflo}). We claim that these learning algorithms, whether explicitly derived as such or not, are all implicitly approximations to RTRL, each a special case of a general class of techniques for compressing ${\bf M}^{(t)}$. In the following section, we clarify how each of these learning algorithms fits into this broad structure.
\subsubsection{Approximations to RTRL}\label{rtrl_approx}
To concretely illuminate these ideas, we will work with a special case of $F_{{\bf w}}$, a time-continuous vanilla RNN:
\begin{equation}
{\bf a}^{(t)} = F_{{\bf w}}({\bf a}^{(t-1)}, {\bf x}^{(t)}) = (1 - \alpha) {\bf a}^{(t-1)} + \alpha\phi({\bf W}{\bf \hat{a}}^{(t-1)}),
\label{vanilla_rnn}
\end{equation}
where ${\bf \hat{a}}^{(t-1)} = \text{concat}({\bf a}^{(t-1)}, {\bf x}^{(t)}, 1) \in \mathbb{R}^m$, ${\bf W} \in \mathbb{R}^{n \times m}$, $\phi: \mathbb{R}^n \rightarrow \mathbb{R}^n$ is some point-wise nonlinearity (e.g. $\tanh$), and $\alpha \in (0, 1]$ is the network's inverse time constant. The trainable parameters $w_p$ are folded via the indexing $p = i\times n + j$ into the weight matrix $W_{ij}$, whose columns hold the recurrent weights, the input weights, and a bias. By reshaping $w_{p}$ into its natural matrix form $W_{ij}$, we can write the influence matrix as an order-3 {\bf influence tensor}
\begin{equation*}
M^{(t)}_{kij} = \partial a^{(t)}_k / \partial W_{ij}.
\end{equation*}
Thus $M^{(t)}_{kij}$ specifies the effect on the $k$-th unit of perturbing the direct connection from the $j$-th unit to the $i$-th unit. The immediate influence can also be written as a tensor. By differentiating Eq.~\eqref{vanilla_rnn}, we see it takes the sparse form
\begin{equation*}
\overline{M}^{(t)}_{kij} = \partial a^{(t)}_k / \partial W^{(t)}_{ij} = \alpha \delta_{ki}\phi'(h^{(t)}_i) \hat{a}^{(t-1)}_j,
\end{equation*}
because $W_{ij}$ can affect the $k$-th unit directly only if $k = i$. Many approximations of RTRL involve a decomposition of $M_{kij}^{(t)}$ into a product of lower-order tensors. For example, UORO represents $M_{kij}^{(t)}$ by an outer product $A_k^{(t)} B_{ij}^{(t)}$, which has a memory requirement of only $\mathcal{O}(n^2)$. Similarly, KF-RTRL uses a Kronecker-product decomposition $A_j^{(t)} B_{ki}^{(t)}$. We can generalize these cases into a set of six possible decompositions of $M^{(t)}_{kij}$ into products of lower-order tensors $A^{(t)}$ and $B^{(t)}$:
\begin{equation*}
M_{kij}^{(t)} \approx \begin{cases}
A_{k}^{(t)} B_{ij}^{(t)} & \text{UORO}, \S\ref{uoro}\\
A_{j}^{(t)} B_{ki}^{(t)} & \text{KF-RTRL}, \S\ref{kf-rtrl}\\
A_{i}^{(t)} B_{kj}^{(t)} & \text{``Reverse" KF-RTRL}, \S\ref{ikf-rtrl}\\
A_{ki}^{(t)} B_{ij}^{(t)} & \text{KeRNL/RFLO}, \S\ref{kernl}/ \S\ref{rflo}\\
A_{kj}^{(t)} B_{ij}^{(t)} & \text{Unexplored}\\
A_{ki}^{(t)} B_{kj}^{(t)} & \text{Unexplored}
\end{cases}.
\label{rtrl_decompositions}
\end{equation*}
Each such decomposition has a memory requirement of $\mathcal{O}(n^2)$. Of course, it is not sufficient to write down an idealized decomposition for a particular time point; there must exist some efficient way to \emph{update} the decomposition as the network runs forwards. We now go through each algorithm and show the mathematical techniques used to derive update equations and categorize them by the criteria outlined in Table~\ref{algtable}.

\subsection{Unbiased Online Recurrent Optimization (UORO)}\label{uoro}
\cite{tallec2017unbiased} discovered a technique for approximating ${\bf M}^{(t)} \in \mathbb{R}^{n \times P}$ as an outer product ${\bf A}^{(t)}{\bf B}^{(t)}$, where ${\bf A}^{(t)} \in \mathbb{R}^{n \times 1}$ and ${\bf B}^{(t)} \in \mathbb{R}^{1 \times P}$. The authors proved a crucial lemma (see \hyperref[sec.lemma]{Appendix A} or \citealp{tallec2017unbiased}) that gives, in closed form, an unbiased rank-1 estimate of a given matrix over the choice of a random vector $\boldsymbol{\nu} \in \mathbb{R}^{n}$ with $\mathbb{E}[\nu_i \nu_j] \propto \delta_{ij}$ and $\mathbb{E}[\nu_i] = 0$. They leverage this result to derive a closed-form update rule for ${\bf A}^{(t)}$ and ${\bf B}^{(t)}$ at each time step, without ever having to explicitly (and expensively) calculate ${\bf M}^{(t)}$. We present an equivalent formulation in terms of tensor components, i.e.,
\begin{equation*}
M^{(t)}_{kij} \approx A^{(t)}_k B^{(t)}_{ij},
\end{equation*}
where $B^{(t)}_{ij}$ represents the ``rolled-up" components of ${\bf B}^{(t)}$, as in $W_{ij}$ w.r.t. ${\bf w}$. Intuitively, the $kij$-th component of the influence matrix is constrained to be the product of the $k$-th unit's ``sensitivity" $A^{(t)}_k$ and the $ij$-th parameter's ``efficacy" $B^{(t)}_{ij}$. Eqs. \eqref{uoro_update} and \eqref{uoro_unbiased} show the form of the update and why it is unbiased over $\boldsymbol{\nu}$, respectively:\\
\begin{align*}
A^{(t)}_k B^{(t)}_{ij} &= \left(\rho_0 \sum_{k'} J^{(t)}_{kk'} A^{(t-1)}_{k'} + \rho_1 \nu_k\right)\left(\rho_0^{-1} B^{(t-1)}_{ij} + \rho_1^{-1}\sum_{k'}\nu_{k'} \overline{M}^{(t)}_{k'ij}\right) \\
&= \sum_{k'} J^{(t)}_{kk'} A^{(t-1)}_{k'}B^{(t-1)}_{ij} + \sum_{k'} \nu_k \nu_{k'} \overline{M}^{(t)}_{k'ij}\\
&+ \sum_{k'} \nu_{k'} \left[\rho_1 \rho_0^{-1}\delta_{kk'} B^{(t-1)}_{ij} + \rho_0 \rho_1^{-1}\overline{M}^{(t)}_{k'ij}\sum_{k''}J^{(t)}_{k'k''}A^{(t-1)}_{k''}\right]\numberthis \label{uoro_update}\\
\implies \mathbb{E}\left[A^{(t)}_k B^{(t)}_{ij}\right] &= \sum_{k'} J^{(t)}_{kk'} \mathbb{E}\left[A^{(t-1)}_{k'}B^{(t-1)}_{ij}\right] + \sum_{k'}\mathbb{E}[\nu_k \nu_{k'}]\overline{M}^{(t)}_{k'ij}\\
&+ \sum_{k'} \mathbb{E}[\nu_{k'}]\left( \text{cross terms} \right)\\
&= \sum_{k'} J^{(t)}_{kk'} M^{(t-1)}_{k'ij} + \sum_{k'} \delta_{kk'}\overline{M}^{(t)}_{k'ij} + \sum_{k'} 0 \times (\text{cross terms})\\
&= \sum_{k'} J^{(t)}_{kk'} M^{(t-1)}_{k'ij} + \overline{M}^{(t)}_{kij}\\
&= M^{(t)}_{kij}. \numberthis \label{uoro_unbiased}
\end{align*}
The cross terms vanish in expectation because $\mathbb{E}[\nu_k] = 0$. Thus, by induction over $t$, the estimate of $M^{(t)}_{kij}$ remains unbiased at every time step. The constants $\rho_0, \rho_1 \in \mathbb{R}^{>0}$ are chosen at each time step to minimize total variance of the estimate by balancing the norms of the cross terms. This algorithm's update is {\bf stochastic} due to its reliance on the random vector $\boldsymbol{\nu}$, but it is in {\bf closed form} because it has an explicit update formula (Eq.~\ref{uoro_update}). Both its memory and computational complexity are $\mathcal{O}(n^2)$.

\subsection{Kronecker-Factored RTRL (KF-RTRL)} \label{kf-rtrl}

\cite{mujika2018approximating} leverage the same lemma as in UORO, but using a decomposition of ${\bf M}^{(t)}$ in terms of a Kronecker product ${\bf A}^{(t)} \otimes {\bf B}^{(t)}$, where now ${\bf A}^{(t)} \in \mathbb{R}^{1 \times m}$ and ${\bf B}^{(t)}~\in~ \mathbb{R}^{n \times n}$. This decomposition is more natural, because the immediate influence ${\bf \overline{M}}^{(t)}$ factors \emph{exactly} as a Kronecker product ${\bf \hat{a}}^{(t)} \otimes {\bf D}^{(t)}$ for vanilla RNNs, where $D^{(t)}_{ki} = \alpha \delta_{ki} \phi'(h^{(t)}_i)$. To derive the update rule for UORO, one must first generate a rank-1 estimate of ${\bf \overline{M}}^{(t)}$ as an intermediate step, introducing more variance, but in KF-RTRL, this step is unnecessary. In terms of components, the compression takes the form
\begin{equation*}
M^{(t)}_{kij} \approx A^{(t)}_j B^{(t)}_{ki},
\end{equation*}
which is similar to UORO, modulo a cyclic permutation of the indices. Given a sample $\boldsymbol{\nu} \in \mathbb{R}^2$ of only 2 i.i.d. random variables, again with $\mathbb{E}[\nu_i \nu_j] \propto \delta_{ij}$ and $\mathbb{E}[\nu_i] = 0$, the update takes the form shown in Eqs. \eqref{kfrtrl_a_update} and \eqref{kfrtrl_b_update}:
\begin{align*}
A^{(t)}_j &= \left(\nu_0 \rho_0 A^{(t-1)}_j + \nu_1 \rho_1 \hat{a}^{(t-1)}_j\right) \numberthis \label{kfrtrl_a_update}\\
B^{(t)}_{ki} &= \left(\nu_0 \rho_0^{-1} \sum_{k'} J^{(t)}_{kk'} B^{(t-1)}_{k'i} + \nu_1 \rho_1^{-1}\alpha\delta_{ki} \phi'(h^{(t)}_i)\right) \numberthis \label{kfrtrl_b_update}\\
\implies A^{(t)}_j B^{(t)}_{ki} &= \nu_0^2 \sum_{k'} J^{(t)}_{kk'} A^{(t-1)}_j B^{(t-1)}_{k'i} + \nu_1^2\alpha\delta_{ki} \phi'(h^{(t)}_i)\hat{a}^{(t-1)}_j + \text{cross-terms}
\label{kfrtrl_update}\\
\implies \mathbb{E}\left[A^{(t)}_j B^{(t)}_{ki}\right] &= \sum_{k'} J^{(t)}_{kk'} \mathbb{E}\left[A^{(t-1)}_j B^{(t-1)}_{k'i}\right] + \alpha\delta_{ki} \phi'(h^{(t)}_i)\hat{a}^{(t-1)}_j\\
&= \sum_{kk'} J^{(t)}_{kk'} M^{(t-1)}_{k'ij} + \overline{M}^{(t)}_{kij}\\
&= M^{(t)}_{kij}.
\end{align*}
As in UORO, the cross terms vanish in expectation, and the estimate is unbiased by induction over $t$. This algorithm's updates are also {\bf stochastic} and in {\bf closed form}. Its memory complexity is $\mathcal{O}(n^2)$, but its computation time is $\mathcal{O}(n^3)$ because of the matrix-matrix product in Eq.~\eqref{kfrtrl_b_update}.

\subsection{Reverse KF-RTRL (R-KF-RTRL)} 
\label{ikf-rtrl}

Our exploration of the space of different approximations naturally raises a question: is an approximation of the form
\begin{equation}
M^{(t)}_{kij} \approx A^{(t)}_i B^{(t)}_{kj}
\label{ikfrtrl_approx}
\end{equation}
also possible? We refer to this method as ``Reverse" KF-RTRL (R-KF-RTRL) because, in matrix notation, this would be formulated as ${\bf M}^{(t)} \approx {\bf B}^{(t)} \otimes {\bf A}^{(t)}$, where ${\bf A}^{(t)} \in \mathbb{R}^{1 \times n}$ and ${\bf B}^{(t)} \in \mathbb{R}^{n \times m}$. We propose the following update for $A^{(t)}_i$ and $B^{(t)}_{kj}$ in terms of a random vector $\boldsymbol{\nu} \in \mathbb{R}^n$:
\begin{align*}
A^{(t)}_i B^{(t)}_{kj} &= \left(\rho_0 A^{(t-1)}_i + \rho_1\nu_i\right)\left(\rho_0^{-1} \sum_{k'} J^{(t)}_{kk'}B^{(t-1)}_{k'j}  + \rho_1^{-1}\sum_{i'} \nu_{i'} \overline{M}^{(t)}_{ki'j}\right) \numberthis \label{ikfrtrl_update}\\
&= \sum_{k'} J^{(t)}_{kk'}A^{(t-1)}_i B^{(t-1)}_{k'j} + \sum_{i'} \nu_i \nu_{i'} \overline{M}^{(t)}_{ki'j} + \text{cross-terms}\\
\implies \mathbb{E}\left[A^{(t)}_i B^{(t)}_{kj}\right] &= \sum_{k'} J^{(t)}_{kk'}\mathbb{E}\left[A^{(t-1)}_i B^{(t-1)}_{k'j}\right] + \overline{M}^{(t)}_{kij}\\
&= \sum_{k'} J^{(t)}_{kk'} M^{(t-1)}_{k'ij} + \overline{M}^{(t)}_{kij}\\
&= M^{(t)}_{kij}. \numberthis \label{ikfrtrl_unbiased}
\end{align*}
Eq.~\eqref{ikfrtrl_unbiased} shows that this estimate is unbiased, using updates that are {\bf stochastic} and in {\bf closed form}, like its sibling algorithms. Its memory and computational complexity are $\mathcal{O}(n^2)$ and $\mathcal{O}(n^3)$, respectively. R-KF-RTRL is actually more similar to UORO than KF-RTRL, because $\overline{M}^{(t)}_{kij}$ does not naturally factor like Eq.~\eqref{ikfrtrl_approx}, introducing more variance. Worse, it has the computational complexity of KF-RTRL due to the matrix-matrix multiplication in Eq.~\eqref{ikfrtrl_update}. KF-RTRL stands out as the most effective of these 3 algorithms, because it estimates ${\bf M}^{(t)}$ with the lowest variance due to its natural decomposition structure. (See \citealp{mujika2018approximating} for variance calculations.)
\subsubsection{Optimal Kronecker-Sum Approximation (OK)}
We briefly mention an extension of KF-RTRL by \cite{benzing2019optimal}, where the influence matrix is approximated not by 1 but rather a sum of $r$ Kronecker products, or, in components
\begin{equation*}
M^{(t)}_{kij} \approx \sum_{l=1}^r A^{(t)}_{lj} B^{(t)}_{lki}.
\end{equation*}
On the RTRL update, the $k$ index of $B^{(t)}_{lki}$ is propagated forward by the Jacobian, and then the immediate influence---itself a Kronecker product---is added. Now $M^{(t)}_{kij}$ is approximated by $r+1$ Kronecker products
\begin{equation*}
M^{(t)}_{kij} \approx \sum_{l=1}^r A^{(t-1)}_{lj} J^{(t)}_{kk'}B^{(t-1)}_{lk'i} + \alpha \hat{a}^{(t-1)}_j \delta_{ki} \phi'(h^{(t)}_i),
\end{equation*}
but the authors developed a technique to optimally reduce this sum back to $r$ Kronecker products, keeping the memory complexity $\mathcal{O}(rn^2)$ and computational complexity $\mathcal{O}(rn^3)$ constant. This update is {\bf stochastic} because it requires explicit randomness in the flavor of the above algorithms, and it is {\bf numerical} because there is no closed form solution to the update. We leave the details to the original paper.
\subsection{Kernel RNN Learning (KeRNL)} \label{kernl}
\cite{roth2018kernel} developed a learning algorithm for RNNs that is essentially a compression of the influence matrix of the form $M^{(t)}_{kij} \approx A_{ki} B^{(t)}_{ij}$. We will show that this algorithm is also an implicit approximation of RTRL, although the update rules are fundamentally different than those for UORO, KF-RTRL and R-KF-RTRL. The {\bf eligibility trace} ${\bf B}^{(t)} \in \mathbb{R}^{n \times m}$ updates by temporally filtering the immediate influences $\alpha \phi'(h^{(t)}_i) \hat{a}^{(t-1)}_j$ with unit-specific, learnable timescales $\alpha_i$:
\begin{equation}
B^{(t)}_{ij} = (1 - \alpha_i)B^{(t-1)}_{ij} + \alpha \phi'(h^{(t)}_i) \hat{a}^{(t-1)}_j.
\label{e_trace}
\end{equation}
The {\bf sensitivity matrix} ${\bf A} \in \mathbb{R}^{n \times n}$ is chosen to approximate the multi-step Jacobian $\partial a^{(t)}_k / \partial a^{(t')}_i$ with help from the learned timescales:
\begin{equation}
\frac{\partial a^{(t)}_k}{\partial a^{(t')}_i} \approx A_{ki} (1 - \alpha_i)^{(t - t')}.
\label{sensitivity_approx}
\end{equation}
We will describe how ${\bf A}$ is learned later, but for now we assume this approximation holds and use it to show how the KeRNL update is equivalent to that of RTRL. We have dropped the explicit time-dependence from ${\bf A}$, because it updates too slowly for Eq.~\eqref{sensitivity_approx} to be specific to any one time point. If we unpack this approximation by one time step, we uncover the consistency relation
\begin{equation}
A_{ki}(1 - \alpha_i) \approx \sum_{k'} J^{(t)}_{kk'} A_{k'i}.
\label{A_consistency}
\end{equation}
By taking $t=t'$ in Eq.~\eqref{sensitivity_approx} and rearranging Eq.~\eqref{A_consistency}, we see this approximation implicitly assumes both
\begin{equation}
A_{ki} \approx
\begin{cases}
\delta_{ki}\\
(1 - \alpha_i)^{-1}\sum_{k'}J^{(t)}_{kk'}A_{k'i}
\end{cases}.
\label{A_kernl_approx}
\end{equation}
Then the eligibility trace update effectively implements the RTRL update, assuming inductively that $M^{(t-1)}_{kij}$ is well approximated by $A_{ki}B^{(t-1)}_{ij}$:
\begin{align*}
A_{ki}B^{(t)}_{ij} &= A_{ki}\left[(1 - \alpha_i)B^{(t-1)}_{ij} + \alpha \phi'(h^{(t)}_i) \hat{a}^{(t-1)}_j\right]\\ &= A_{ki} (1 - \alpha_i) B^{(t-1)}_{ij} +  \alpha A_{ki}\phi'(h^{(t)}_i) \hat{a}^{(t-1)}_j\\
&\approx \sum_{k'} J^{(t)}_{kk'} A_{k'i} B^{(t-1)}_{ij} + \alpha \delta_{ki} \phi'(h^{(t)}_i) \hat{a}^{(t-1)}_j \numberthis \label{use_A_approx}\\ 
&= \sum_{kk'} J^{(t)}_{kk'} M^{(t-1)}_{k'ij} + \overline{M}^{(t)}_{kij}\\
&= M^{(t)}_{kij}.
\end{align*}
In Eq.~\eqref{use_A_approx}, we use each of the special cases from Eq.~\eqref{A_kernl_approx}. Of course, the $A_{ki}$ and $\alpha_i$ have to be learned, and \cite{roth2018kernel} use gradient descent to do so. We leave details to the original paper; briefly, they run in parallel a perturbed forward trajectory to estimate the LHS of Eq.~\eqref{sensitivity_approx} and then perform SGD on the squared difference between the LHS and RHS, giving gradients for $A_{ki}$ and $\alpha_i$.

KeRNL uses {\bf deterministic} updates because it does not need explicit random variables. While the $B^{(t)}_{ij}$ update is in closed form via Eq.~\eqref{e_trace}, the updates for $A_{ki}$ and $\alpha_i$ are {\bf numerical} because of the need for SGD to train them to obey Eq.~\eqref{sensitivity_approx}. Both its memory and computational complexities are $\mathcal{O}(n^2)$.
\subsection{Random-Feedback Online Learning (RFLO)} \label{rflo}
\begin{figure}[t]
    \centering
    \includegraphics[width=\textwidth]{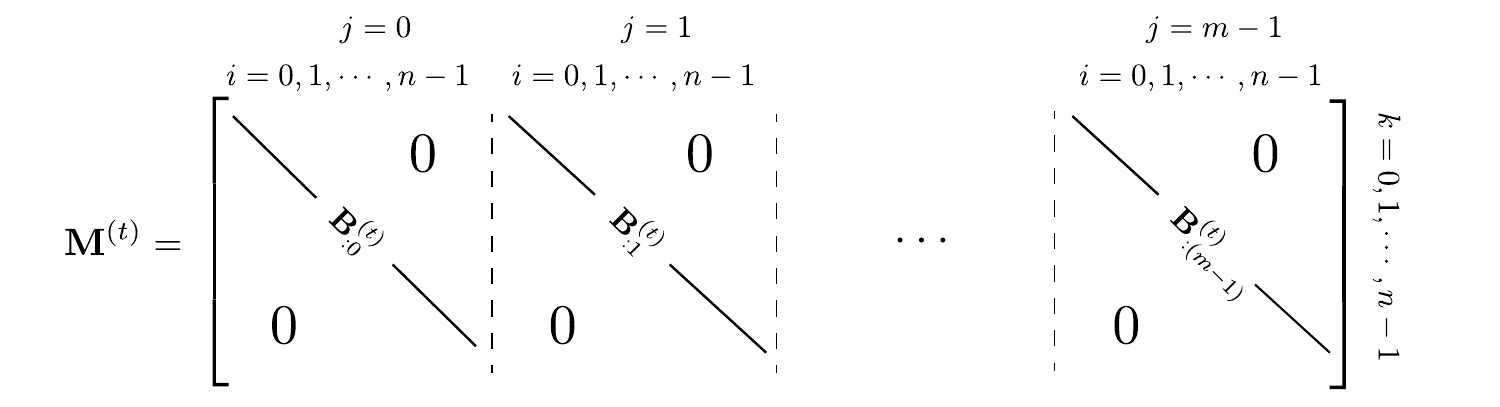}
    \caption{A visualization of the influence matrix and its 3 indices $k$, $i$, and $j$. In RFLO, the filtered immediate influences, stored in $B^{(t)}_{ij}$, sparsely populate the influence matrix along the diagonals.}
    \label{fig:rflo_M}
\end{figure}

Coming from a computational neuroscience perspective, \cite{murray2019local} developed a beautifully simple and biologically plausible learning rule for RNNs, which he calls Random-Feedback Online Learning (RFLO). He formulates the rule in terms of an eligibility trace $B^{(t)}_{ij}$ that filters the non-zero immediate influence elements $\phi'(h^{(t)}_i) \hat{a}^{(t-1)}_j$ by the network inverse time constant $\alpha$:
\begin{equation*}
B^{(t)}_{ij} = (1 - \alpha)B^{(t-1)}_{ij} + \alpha \phi'(h^{(t)}_i) \hat{a}^{(t-1)}_j.
\label{rflo_e_trace}
\end{equation*}
Then the approximate gradient is ultimately calculated\footnote{As the ``random feedback" part of the name suggests, Murray goes a step further in approximating $\overline{c}^{(t)}_k$ by random feedback weights \'a la \citealp{lillicrap2016random}, but we assume exact feedback in this paper for easier comparisons with other algorithms.} as
\begin{equation*}
\fracpartial{L^{(t)}}{W_{ij}} \approx \overline{c}^{(t)}_i B^{(t)}_{ij}.
\end{equation*}
By observing that
\begin{equation*}
\overline{c}^{(t)}_i B^{(t)}_{ij} = \sum_k \overline{c}^{(t)}_k \delta_{ki} B^{(t)}_{ij},
\end{equation*}
we see that RFLO is a special case of KeRNL, in which we fix $A_{ki} = \delta_{ki}$, $\alpha_i = \alpha$. Alternatively, and as hinted in the original paper, we can view RFLO as a special case of RTRL under the approximation $J^{(t)}_{kk'} \approx (1 - \alpha) \delta_{kk'}$, because the RTRL update reduces to RFLO with $M^{(t)}_{kij} = \delta_{ki}B^{(t)}_{ij}$ containing $B^{(t)}_{ij}$ along the diagonals:
\begin{align*}
M^{(t)}_{kij} &= \sum_{k'}J^{(t)}_{kk'}M^{(t-1)}_{k'ij} + \overline{M}^{(t)}_{kij}\\
&= (1 - \alpha) \sum_{k'}\delta_{kk'}M^{(t-1)}_{k'ij}+ \overline{M}^{(t)}_{kij}\\
&= (1 - \alpha)M^{(t-1)}_{kij} + \alpha \delta_{ki}\phi'(h^{(t)}_i) \hat{a}^{(t-1)}_j. \numberthis
\end{align*}
Fig.~\ref{fig:rflo_M} illustrates how ${\bf B}^{(t)}$ is contained in the influence matrix ${\bf M}^{(t)}$. This algorithm's update is {\bf deterministic} and in {\bf closed form}, with memory and computational complexity $\mathcal{O}(n^2)$.

\section{Future-facing algorithms}
\subsection{Backpropagation Through Time (BPTT)}
For many applications, a recurrent network is unrolled only for some finite number of time steps, and backpropagation through time (BPTT) manifests as the computation of the sum $\partial L^{(t)} / \partial {\bf w}^{(s)}$ over every $s \leq t$ in the graph. This can be efficiently accomplished using
\begin{equation}
{\bf c}^{(t)} = {\bf \overline{c}}^{(t)} + {\bf J}^{(t+1)} {\bf c}^{(t+1)}
\label{ff_again}
\end{equation}
(see Eq.~\ref{ff_relation}) to propagate credit assignment backwards. However, in our framework, where a network is run on an infinite-time horizon, there are two qualitatively different ways of unrolling the network. We call them ``efficient" and ``future-facing" BPTT.
\subsubsection{Efficient backpropagation through time (E-BPTT)}\label{e-bptt}
\begin{figure}[ht]
    \centering
    \includegraphics{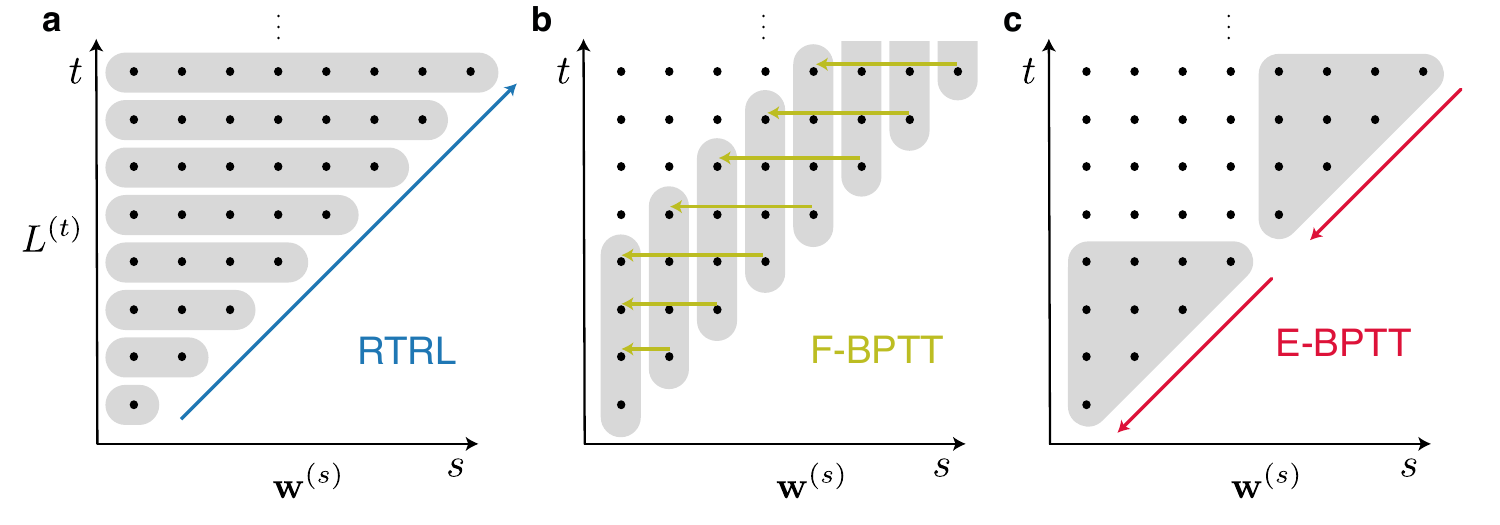}
    \caption{A visualization of various exact gradient methods. Each plot contains a lattice of points, representing derivatives $\partial L^{(t)} / \partial {\bf w}^{(s)}$ for $s \leq t$, with gray boxes representing individual gradients passed to the optimizer. {\bf a)} RTRL sums these derivatives into gradients for fixed $t$, using the PF relation (Eq.~\ref{pf_relation}, \S\ref{pf_ff_online_learning}) to efficiently derive successive gradients (blue arrow). {\bf b)} F-BPTT sums these derivatives into gradients for fixed $s$ by backpropagating through time (yellow arrows). {\bf c)} E-BPTT creates a triangular gradient for non-overlapping subgraphs, using the FF relation (Eq.~\ref{ff_relation}, \S\ref{pf_ff_online_learning}) for efficient computation (red arrows). Here, the truncation horizon is $T = 4$.}
    \label{fig:triangles}
\end{figure}

For this method, we simply divide the graph into non-overlapping segments of truncation length $T$ and perform BPTT between $t-T$ and $t$ as described above, using Eq.~\eqref{ff_again}. It takes $\mathcal{O}(n^2 T)$ computation time to compute one gradient, but since this computation is only performed once every $T$ time steps, the computation time is effectively $\mathcal{O}(n^2)$, with memory requirement $\mathcal{O}(nT)$. A problem with this approach is that it does not treat all time points the same: an application of ${\bf w}$ occurring near the end of the graph segment has less of its future influence accounted for than applications of ${\bf w}$ occurring before it, as can be visualized in Fig.~\ref{fig:triangles}. And since any one gradient passed to the optimizer is a sum across both $t$ and $s$, it is not an online algorithm by the framework we presented in \S\ref{pf_ff_online_learning}. Therefore, for the purpose of comparing with online algorithms, we also show an alternative version of BPTT that calculates a future-facing gradient (up to truncation) $\partial \mathcal{L} / \partial {\bf w}^{(t)}$ for every $t$.
\subsubsection{Future-facing backpropagation through time (F-BPTT)}
In this version of BPTT, we keep a dynamic list of truncated credit assignment estimates ${\bf \hat{c}}^{(s)}$ for times $s = t - T, \cdots, t - 1$:
\begin{equation*}
\left[{\bf \hat{c}}^{(t-T)}, \cdots, {\bf \hat{c}}^{(t-1)}\right],
\end{equation*}
where each truncated credit assignment estimate includes the influences of ${\bf a}^{(s)}$ only up to time $t-1$:
\begin{equation*}
{\bf \hat{c}}^{(s)} = \sum_{t' = s}^{t-1} \fracpartial{L^{(t')}}{{\bf a}^{(s)}}.
\end{equation*}
At current time $t$, every element ${\bf \hat{c}}^{(s)}$ is extended by adding $\partial L^{(t)} / \partial {\bf a}^{(s)}$, calculated by backpropagating from the current loss $L^{(t)}$, while the explicit credit assignment ${\bf \overline{c}}^{(t)}$ is appended to the front of the list. To compensate, the oldest credit assignment estimate ${\bf \hat{c}}^{(t-T)}$ is removed and combined with the immediate influence to form a (truncated) gradient
\begin{equation*}
{\bf \hat{c}}^{(t-T)}{\bf \overline{M}}^{(t-T)} = \sum_{t' = t-T}^{t} \fracpartial{L^{(t')}}{{\bf a}^{(t-T)}}\fracpartial{{\bf a}^{(t-T)}}{{\bf w}^{(t-T)}} = \sum_{t' = t-T}^{t} \fracpartial{L^{(t')}}{{\bf w}^{(t-T)}} \approx \fracpartial{\mathcal{L}}{{\bf w}^{(t-T)}},
\end{equation*}
which is passed to the optimizer to update the network. This algorithm is ``online" in that it produces strictly future-facing gradients at each time step, albeit delayed by the truncation time $T$ and requiring memory of the network states from $t-T$. Each update step requires $\mathcal{O}(n^2 T)$ computation, but since the update is performed at every time step, computation remains a factor of $T$ more expensive than E-BPTT. Memory requirement is still $\mathcal{O}(nT)$. Fig.~\ref{fig:triangles} illustrates the differences among these methods and RTRL, using a triangular lattice as a visualization tool. Each point in the lattice is one derivative $\partial L^{(t)} / \partial {\bf w}^{(s)}$ with $t \geq s$, and the points are grouped together into discrete gradients passed to the optimizer.
\subsection{Decoupled Neural Interfaces (DNI)} \label{dni}
\cite{jaderberg2017decoupled} developed a framework for online learning by {\bf predicting} credit assignment. Whereas PF algorithms face the problem of a large influence tensor $M^{(t)}_{kij}$ that needs a compressed representation, FF algorithms face the problem of incomplete information: at time $t$, it is impossible to calculate ${\bf c}^{(t)}$ without access to future network variables. The approach of Decoupled Neural Interfaces (DNI) is to simply make a linear prediction of ${\bf c}^{(t)}$ \citep{czarnecki2017understanding} based on the current hidden state ${\bf a}^{(t)}$ and the current labels ${\bf y}^{*(t)}$:
\begin{equation*}
c^{(t)}_i \approx \sum_l \tilde{a}^{(t)}_l A_{li},
\label{dni_approx}
\end{equation*}
where ${\bf \tilde{a}}^{(t)} = \text{concat}({\bf a}^{(t)}, {\bf y}^{*(t)}, 1) \in \mathbb{R}^{m'}$, $m' = n + n_{\text{out}} + 1$, and $A_{li}$ are the components of a matrix ${\bf A} \in \mathbb{R}^{m \times n}$, which parameterizes what the authors call the {\bf synthetic gradient} function. The parameters $A_{li}$ are trained to minimize the loss
\begin{equation}
L^{(t)}_{\text{SG}} = \frac{1}{2}\norm{\sum_l \tilde{a}^{(t)}_l A_{li} - c^{(t)}_i}^2
\label{sg_loss}
\end{equation}
via gradient descent, similar to KeRNL's treatment of $A_{ki}$ and $\alpha_i$ (and we drop the time dependence of $A_{li}$ for the same reason). Of course, this begs the question---the whole point is to avoid calculating ${\bf c}^{(t)}$ explicitly, but calculating the error in Eq.~\eqref{sg_loss} requires access to ${\bf c}^{(t)}$.  So the authors propose a ``bootstrapping" technique analogous to the Bellman equation in Reinforcement Learnin \citep{sutton2018reinforcement}. If we take the FF relation we derived in Eq.~\eqref{ff_relation}
\begin{equation}
{\bf c}^{(t)} = {\bf \overline{c}}^{(t)} + {\bf c}^{(t+1)}{\bf J}^{(t+1)}
\label{ff_relation_again}
\end{equation}
and approximate the appearance of ${\bf c}^{(t+1)}$ with the synthetic gradient estimate ${\bf \tilde{a}}^{(t+1)}{\bf A}$, then Eq.~\eqref{ff_relation_again} provides an estimate of $c^{(t)}_i$ to use in Eq.~\eqref{sg_loss}. Then the update for ${\bf A}$ can be written as
\begin{equation}
\Delta A_{li} \propto - \tilde{a}^{(t)}_l \left[\sum_{l'} \tilde{a}^{(t)}_{l'} A_{l'i} - \left(\overline{c}^{(t)}_i + \sum _{m}\sum_{l'} \tilde{a}^{(t+1)}_{l'} A_{l'm} J^{(t+1)}_{mi}\right)\right]
\label{sg_update}
\end{equation}
with learning rate chosen as a hyperparameter. As in Eq.~\eqref{ff_gradient}, the gradient is calculated by combining the estimated credit assignment for the $i$-th unit with the explicit influence by the $ij$-th parameter:
\begin{equation*}
\fracpartial{\mathcal{L}}{W^{(t)}_{ij}} = \fracpartial{\mathcal{L}}{a^{(t)}_{i}}\fracpartial{a^{(t)}_i}{W^{(t)}_{ij}} = \overline{c}^{(t)}_i \phi'(h^{(t)}_i) \hat{a}^{(t-1)}_j \approx \left(\sum_l \tilde{a}^{(t)}_l A_{li}\right)\phi'(h^{(t)}_i) \hat{a}^{(t-1)}_j
\end{equation*}
This algorithm is {\bf future facing} because it ultimately estimates the effect of applying ${\bf w}$ at current time $t$ on \emph{future} losses. Its updates are {\bf deterministic} and {\bf numerical}, because no explicit randomness is required, but the minimization problem over $A_{li}$ implied by Eq.~\eqref{sg_loss} is approximated by gradient descent rather than solved in closed form. It requires $\mathcal{O}(n^2)$ memory for ${\bf A}$ and $\mathcal{O}(n^2)$ computation for the matrix-vector multiplications in Eq.~\eqref{sg_update}.
\subsubsection{Biological approximation to DNI}
While many of the algorithms we have presented are biologically plausible in the abstract, i.e. temporally/spatially local and requiring no more than $\mathcal{O}(n^2)$ memory, we have not yet discussed any explicit biological implementations. There are a handful of additional considerations for evaluating an algorithm with respect to biological plausibility:
\begin{enumerate}[i]
\itemsep0em
    \item Any equation describing synaptic strength changes (weight updates) must be {\bf local}, i.e. any variables needed to update a synapse connecting the $i$-th and $j$-th units must be physically available to those units.
    \item Matrix-vector multiplication can be implemented by network-wide neural transmission, but input vectors must represent {\bf firing rates} (e.g. post-activations ${\bf a}$) and not membrane potentials (e.g. pre-activations ${\bf h}$), since inter-neuron communication is mediated by spiking.
    \item {\bf Feedback weights} used to calculate ${\bf \overline{c}}$ cannot be perfectly symmetric with ${\bf W}^{\text{out}}$, since there is no evidence for biological weight symmetry (see \citealp{lillicrap2016random}).
    \item Matrices (e.g. ${\bf J}$ or ${\bf A}$) must represent a set of synapses, whose strengths are determined by some local update.
\end{enumerate}
With a few modifications, many of the presented algorithms can satisfy these requirements. We briefly illustrate one particular case with DNI, as shown in \cite{Marschall19}. To address (i), the result of the synthetic gradient operation $\sum_l \tilde{a}^{(t)}_l A_{li}$ can be stored in an electrically isolated neural compartment, in a manner similar to biological implementations of feedforward backpropagation \citep{guerguiev2017towards, sacramento2018dendritic}, to allow for local updates to $W_{ij}$. For (ii), simply pass the bootstrapped estimate of ${\bf \overline{c}}^{(t+1)}$ from Eq.~\eqref{sg_update} through the activation function $\phi$ so that it represents a neural firing rate. For (iii), one can use fixed, random feedback weights ${\bf W}^{\text{fb}}$ instead of the output weights to calculate ${\bf \overline{c}}^{(t)}$, as in \cite{lillicrap2016random}. And for (iv), one can train a set of weights $\mathcal{J}_{ij}$ online to approximate the Jacobian by performing SGD on $L_J^{(t)} = \norm{a^{(t)}_i - \sum_j  \mathcal{J}_{ij}a^{(t-1)}_j}^2$, which encodes the error of the linear prediction of the next network state by $\mathcal{J}_{ij}$. The update rule manifests as 
\begin{equation*}
\Delta \mathcal{J}_{ij} \propto - \left(a^{(t)}_i - \sum_{j'}  \mathcal{J}_{ij'}a^{(t-1)}_{j'}\right)a^{(t-1)}_j,
\end{equation*}
essentially a ``perceptron" learning rule, which is local and biologically realistic. Although this approximation brings no traditional computation speed benefits, it offers a plausible mechanism by which a neural circuit can access its own Jacobian for learning purposes. This technique could be applied to any other algorithm discussed in this paper. We refer to this altered version of DNI as DNI(b) in the experiments section.

\section{Experiments}
\label{results_section}
We run a number of experiments to empirically validate our categorizations and compare performance of the algorithms reviewed here, using two different synthetic tasks.

\subsection{Setup}
We implemented every algorithm presented here in a custom NumPy-based Python module.\footnote{Link to public code repository to be included upon acceptance.} In every simulation, we use gradient descent with a learning rate of $10^{-4}$, the fastest learning rate for which all algorithms are able to converge to a stable steady-state performance. We restrict ourselves to using a batch size of 1, because, in an online setting, the network must learn as data arrive in real time. Most algorithms demand additional configuration decisions and hyperparameter choices: the truncation horizon $T$ (F-BPTT), the initial values of the tensors ${\bf A}$ and ${\bf B}$ (all approximations), the initial values of the learned timescales $\alpha_i$ (KeRNL), the distribution from which $\boldsymbol{\nu}$ is sampled for stochastic updates (UORO, KF-RTRL, R-KF-RTRL), and the learning rate for the numerical updates (KeRNL, DNI). For each algorithm, we independently optimize these choices by hand (see \hyperref[appendix:hyperparameters]{Appendix B} for details).

We evaluate each algorithm's ability to learn two different synthetic tasks: an {\bf additive dependencies} task (Add) inspired by \cite{pitis2016recurrent} and a {\bf mimic target RNN} task (Mimic). In both tasks, a stream of i.i.d. Bernoulli inputs ${\bf x}^{(t)} \in \{0, 1\}^{n_{\text{in}}}$ is provided to the RNN. For Add, $n_{\text{in}} = n_{\text{out}} = 1$. The label $y^{*(t)}$ has a baseline value of 0.5 that increases (or decreases) by 0.5 (or 0.25) if $x = 1$ at $t - t_1$ (or $t - t_2$), for specified lags $t_1$ and $t_2$. The longer the lags of the dependencies, the more difficult the task. We choose $t_1 = 6$ and $t_2 = 10$. In the Mimic task, the labels ${\bf y}^{*(t)} \in \mathbb{R}^{n_{\text{out}}}$ are determined by the outputs of a randomly generated, untrained target RNN that is fed the same input stream $\left\{{\bf x}^{(t')}: t' \leq t\right\}$. We use $n_{\text{in}} = n_{\text{out}} = 32$ in Mimic, chosen so that learning ${\bf W}$ is necessary for strong performance.

\begin{figure}[ht]
    \centering
    \includegraphics{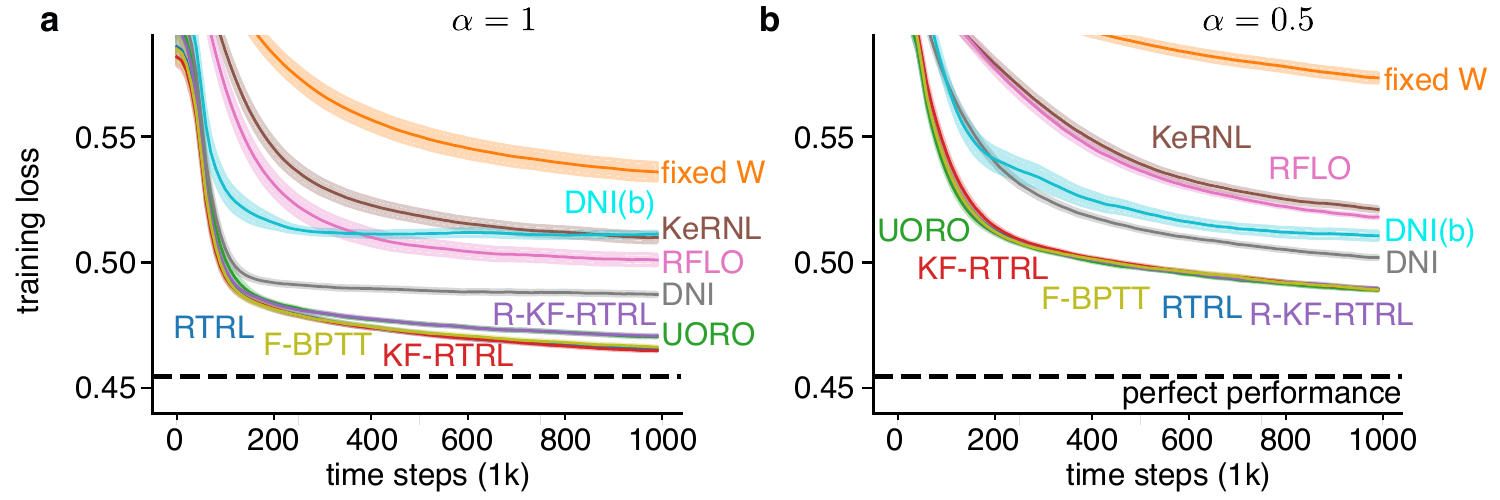}
    \caption{{\bf a)} Cross-entropy loss for networks trained on Add task with $\alpha = 1$ for various algorithms. Lines are means over 20 random seeds (weight initialization and training set generation), and shaded regions represent $\pm 1$ S.E.M. Raw loss curves are first down-sampled by a factor of $10^{-4}$ (rectangular kernel) and then smoothed with a 10-time-step windowed running average. {\bf b)} Same for $\alpha = 0.5$.}
    \label{fig:training_loss_coin}
\end{figure}

For each task, we consider a version on two different time scales: when the network is perfectly discrete ($\alpha = 1$, see Eq.~\ref{vanilla_rnn}) and when the network update has some time continuity ($\alpha = 0.5$). For the $\alpha = 0.5$ case, the tasks are stretched over time by a factor of 2 to compensate, and the dependencies are reduced to $t_1 = 3$ and $t_2 = 5$ in the Add task to keep the difficulty roughly the same.

\subsection{Add task: result and analysis}

Fig.~\ref{fig:training_loss_coin} shows the performance of each learning algorithm on the Add task, in both $\alpha = 1$ and $\alpha = 0.5$ conditions, including a {\bf fixed} ${\bf W}$ algorithm that does not train ${\bf W}$ at all but does train ${\bf W}^{\text{out}}$, as a baseline. As expected, since they compute exact gradients up to truncation, RTRL and F-BPTT perform best, although KF-RTRL is a sufficiently robust approximation to virtually tie with them. R-KF-RTRL and UORO perform similarly and worse than KF-RTRL does, as expected, since these approximations carry significantly more variance than KF-RTRL. However, in the $\alpha = 0.5$ condition, their performance is similar to that of KF-RTRL.

KeRNL is theoretically a stronger approximation of RTRL than RFLO, because of its ability to learn optimal $A_{ki}$ and $\alpha_i$ whereas RFLO has fixed $A_{ki} = \delta_{ki}$ and $\alpha_i = \alpha$. However, the numerical procedure for updating $A_{ki}$ and $\alpha_i$ depends on several configuation/hyperparameter choices. Despite significant effort in exploring this space, we are not able to get KeRNL to perform better than RFLO, suggesting that the procedure for training $A_{ki}$ and $\alpha_i$ does more harm than good. In the original paper, \cite{roth2018kernel} show promising results using RMSProp (\citealp{Tieleman2012}) and batched training, so we suspect that the perturbation-based method for training $A_{ki}$ is simply too noisy for online learning.

DNI sits somewhere between the RFLO-KeRNL cluster and the rest, with its biologically realistic sibling DNI(b) performing slightly worse than DNI, as to be expected, since it is an approximation on top of an approximation. As with KeRNL, DNI's numerical update of $A_{li}$ introduces more hyperparameters and implementation choices, but there is a larger space of configurations in which the updates improve rather than hinder the algorithm's performance.

\begin{figure}[ht]
    \centering
    \includegraphics{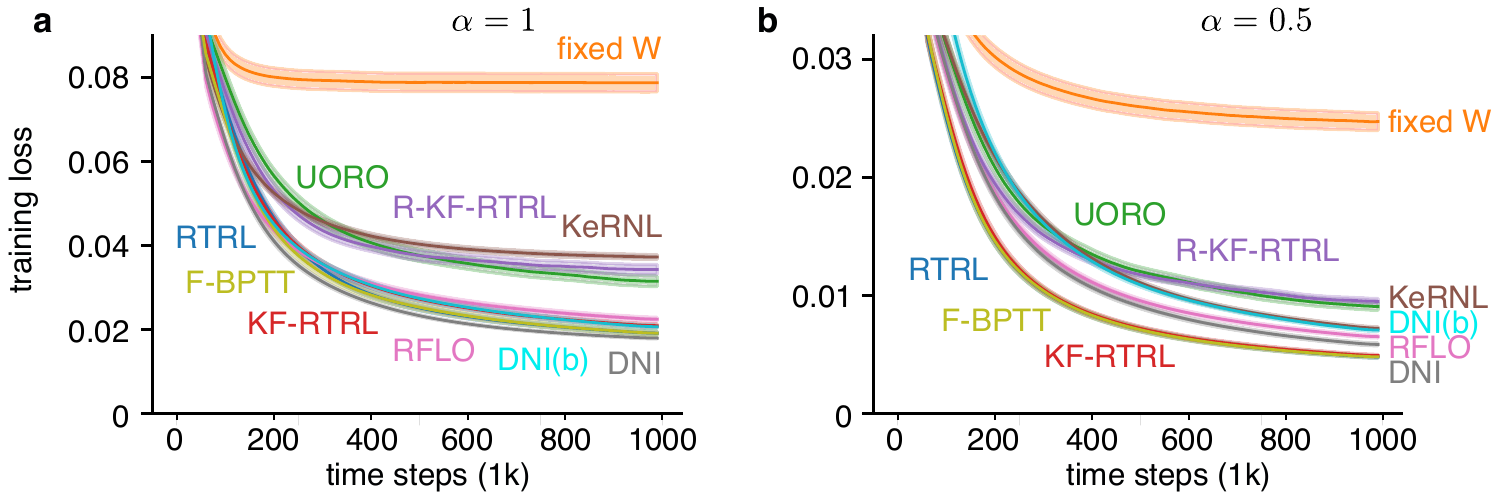}
    \caption{Same as Fig.~\ref{fig:training_loss_coin}, for Mimic task with mean-squared-error loss.}
    \label{fig:training_loss_mimic}
\end{figure}

\subsection{Mimic task: result and analysis}

For the Mimic task (Fig.~\ref{fig:training_loss_mimic}), we see similar clustering of the algorithms but not in the same order. RTRL, F-BPTT, and KF-RTRL perform the best. UORO and R-KF-RTRL perform similarly to each other, but relatively worse than they did on Add. Conversely, DNI, DNI(b), KeRNL, and RFLO perform relatively better on this task than they did on Add. The information content that must be memorized for Add is relatively small ($n_{\text{in}} = 1$), but the time horizon is quite long. In Mimic, it is difficult to quantify the exact time horizon, but it is clear that the network must memorize much more information ($n_{\text{in}} = 32$). Thus perhaps UORO and R-KF-RTRL are effective at maintaining information over time, but the stochasticity in the updates places a limit on how much information can be retained.

\subsection{Gradient similarity analysis}

\begin{figure}[ht]
    \centering
    \includegraphics{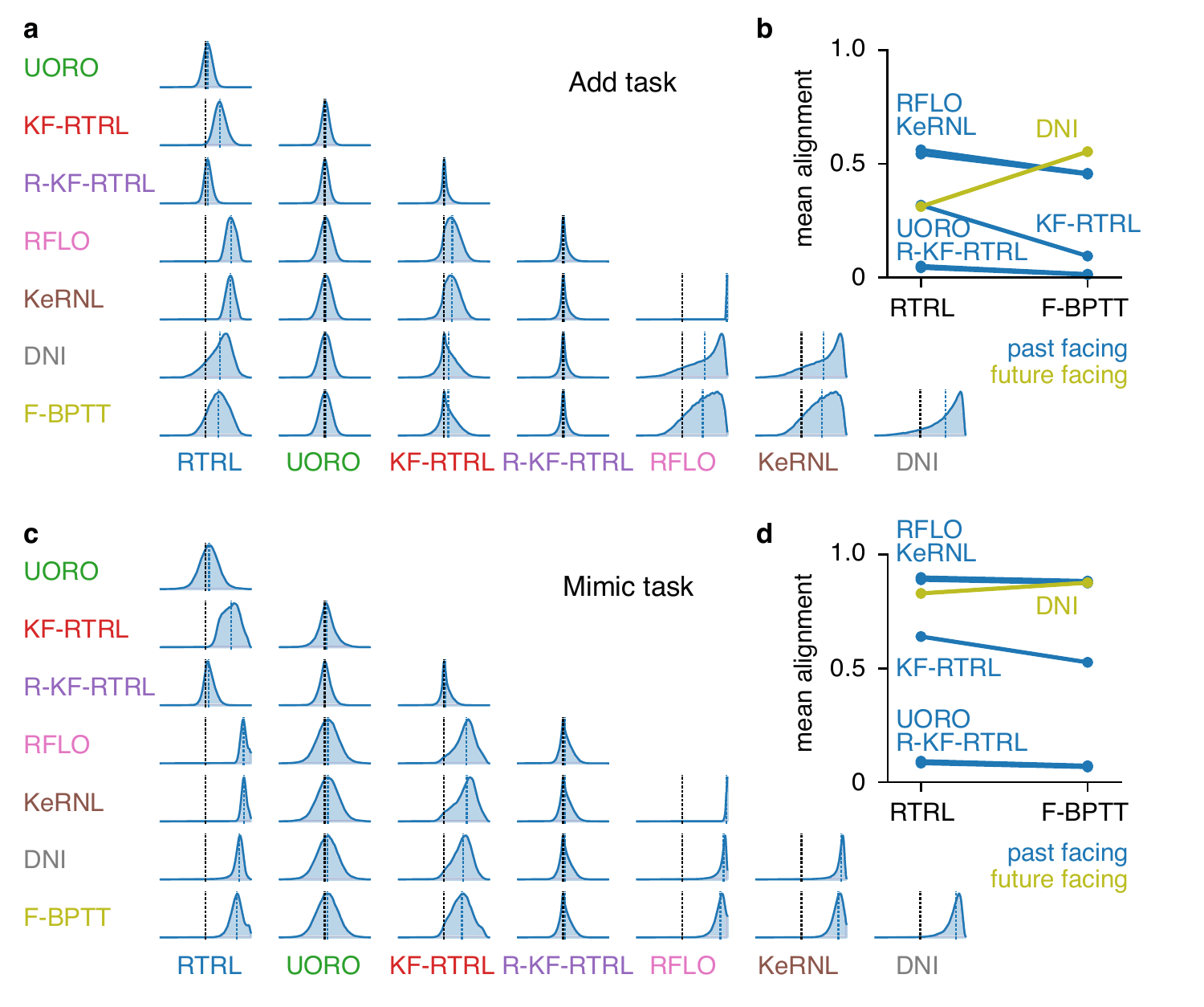}
    \caption{{\bf a)} Histograms of normalized gradient alignments for each pair of algorithms. Gradients are calculated during a simulation of $100k$ time steps of the Add task (same hyperparameters as in Figs.~\ref{fig:training_loss_coin}). Learning follows RTRL gradients, with other algorithms' gradients passively computed for comparison. Mean alignment (dashed blue line) and 0 alignment reference (dashed black line) shown. {\bf b)} Mean alignments of each approximate algorithm with RTRL and F-BPTT, color-coded by past facing (UORO, KF-RTRL, R-KF-RTRL, RFLO, KeRNL) vs.\ future facing (DNI). {\bf c)} Same as (a), for Mimic task. {\bf d)} Same as (b), for Mimic task.}
    \label{fig:alignment_analysis}
\end{figure}

We conduct an in-depth investigation looking beyond task accuracy by directly comparing the gradients (or approximate gradients) produced by each algorithm. Fig.~\ref{fig:alignment_analysis} shows how a given pair of algorithms align on a time-step-by-time-step basis for the Add and Mimic tasks. Each subplot is a histogram giving the distribution of normalized dot products
\begin{equation}
\cos\left(\theta^{(t)}_{XY}\right) = \frac{\Delta^{(t)}_X {\bf W} \cdot \Delta^{(t)}_Y {\bf W}}{\norm{\Delta^{(t)}_X {\bf W}}\norm{\Delta^{(t)}_Y {\bf W}}}
\end{equation}
for (flattened) weight updates $\Delta^{(t)}_X {\bf W}$ and $\Delta^{(t)}_Y {\bf W}$ prescribed by algorithms $X$ and $Y$, respectively, at time $t$.
Figs.~\ref{fig:alignment_analysis}~a,c show qualitatively similar trends:
\begin{enumerate}[i]
    \item As shown directly in Figs.~\ref{fig:alignment_analysis}b,d, PF algorithms (UORO, KF-RTRL, R-KF-RTRL, RFLO, KeRNL) align better with RTRL than with F-BPTT, and vice versa for FF algorithms (DNI).
    \item The deterministic PF algorithms (RFLO and KeRNL) align better with RTRL than the stochastic algorithms (UORO, KF-RTRL, and R-KF-RTRL) align with RTRL.
    \item RFLO and KeRNL align more strongly with each other than any other pair.
    \item UORO and R-KF-RTRL do not align strongly with any other algorithms, despite their ability to train the RNN effectively. UORO's mean alignments with RTRL are 0.043 on Add and 0.084 on Mimic, while R-KF-RTRL's mean alignments with RTRL are 0.050 on Add and 0.092 on Mimic (Fig.~\ref{fig:alignment_analysis}b,d), which are much lower than all other approximate algorithms, even those that perform worse on the task in some cases.
\end{enumerate}
Observations (i)-(iii) validate our categorizations, as similarity according to the normalized alignment corresponds to similarity by the past-facing/future-facing, tensor structure, and stochastic/deterministic criteria. Observation (iv) is puzzling, as it shows that angular alignment with exact algorithms is not predictive of learning performance. 

\begin{figure}[ht]
    \centering
    \includegraphics{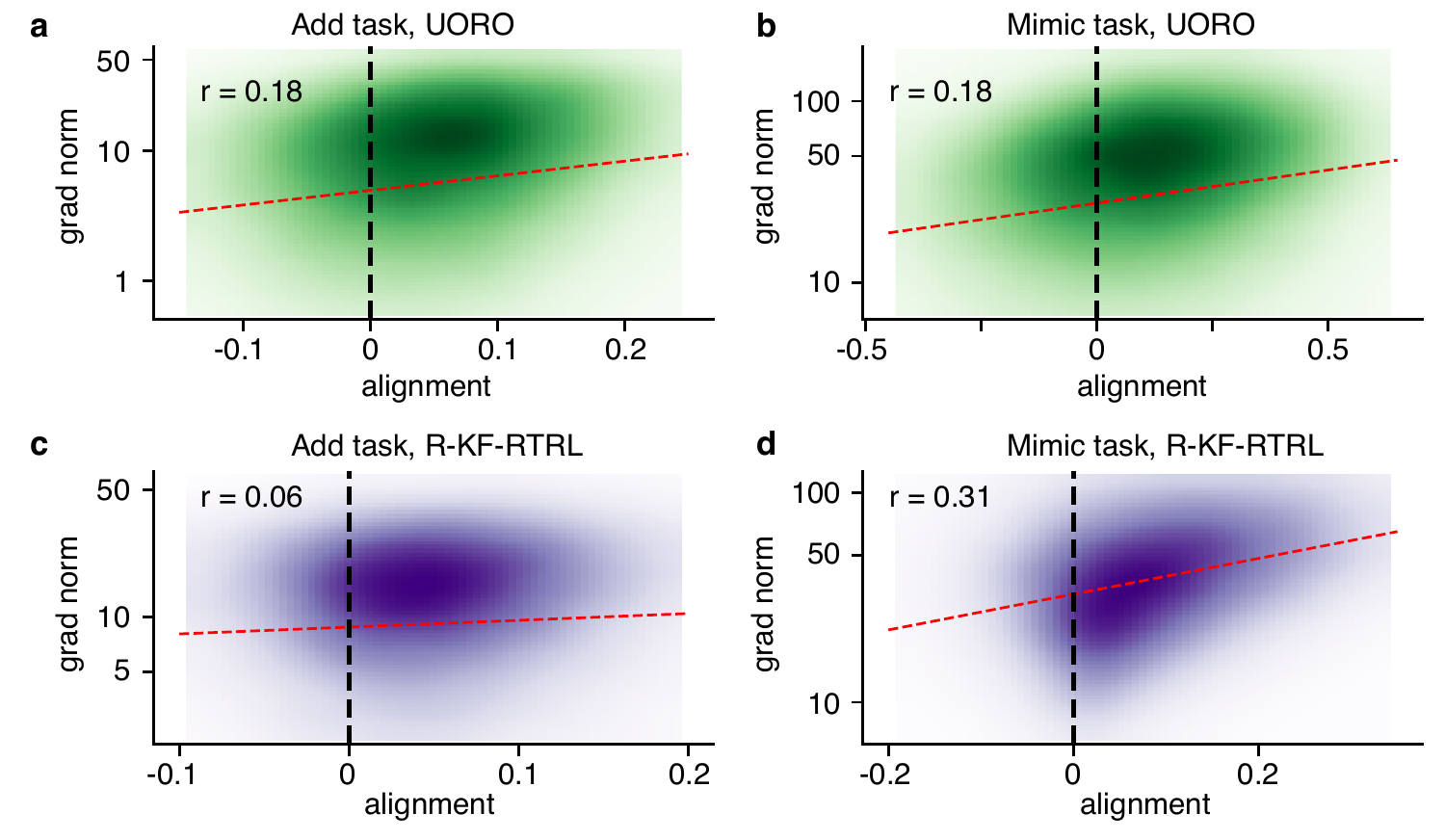}
    \caption{The joint distribution of normalized alignments and gradient norms (log scale). {\bf a)} UORO on Add task, {\bf b)} UORO on Mimic task, {\bf c)} R-KF-RTRL on Add task, {\bf d)} R-KF-RTRL on Mimic task. Color intensity represents (smoothed) observed frequency, based on a sample of 100k time steps, with least-squares regression line in red. The estimated correlation coefficients $r$ are all significant.}
    \label{fig:norm_alignments}
\end{figure}

How are UORO and R-KF-RTRL able to learn at all if their gradients are almost orthogonal with RTRL on average? We address this question for both UORO and R-KF-RTRL by examining the joint distribution of the gradient's alignment with RTRL and the gradient's norm (Fig.~\ref{fig:norm_alignments}). All 4 cases show a statistically significant positive linear correlation between the normalized alignment and the common log of the gradient norm. This observation may partially explain (iv), because larger weight updates occur when UORO happens to strongly align with RTRL. However, these correlations are fairly weak even if statistically significant, and we argue that better algorithm similarity metrics are needed to account for observed differences in performance.


\subsection{RFLO analysis}
\label{rflo_results}

Among all approximate algorithms, RFLO stands out as having the simplest tensor structure and update rule, and it has been empirically demonstrated to be able to train RNNs on long-term dependencies. This is such a severe approximation of RTRL, yet it works so well in practice---and there is no clear understanding why. Although \cite{murray2019local} goes into detail showing how loss decreases despite the random feedback used to approximately calculate ${\bf \overline{c}}^{(t)}$, he does not address the more basic mystery of how RFLO is able to learn despite the significant approximation ${\bf J}^{(t)} \approx (1 - \alpha) {\bf I}$. In this section, we provide some intuition for how this simple learning rule is so successful and empirically validate our claims.

\begin{figure}[ht]
    \centering
    \includegraphics[width=\textwidth]{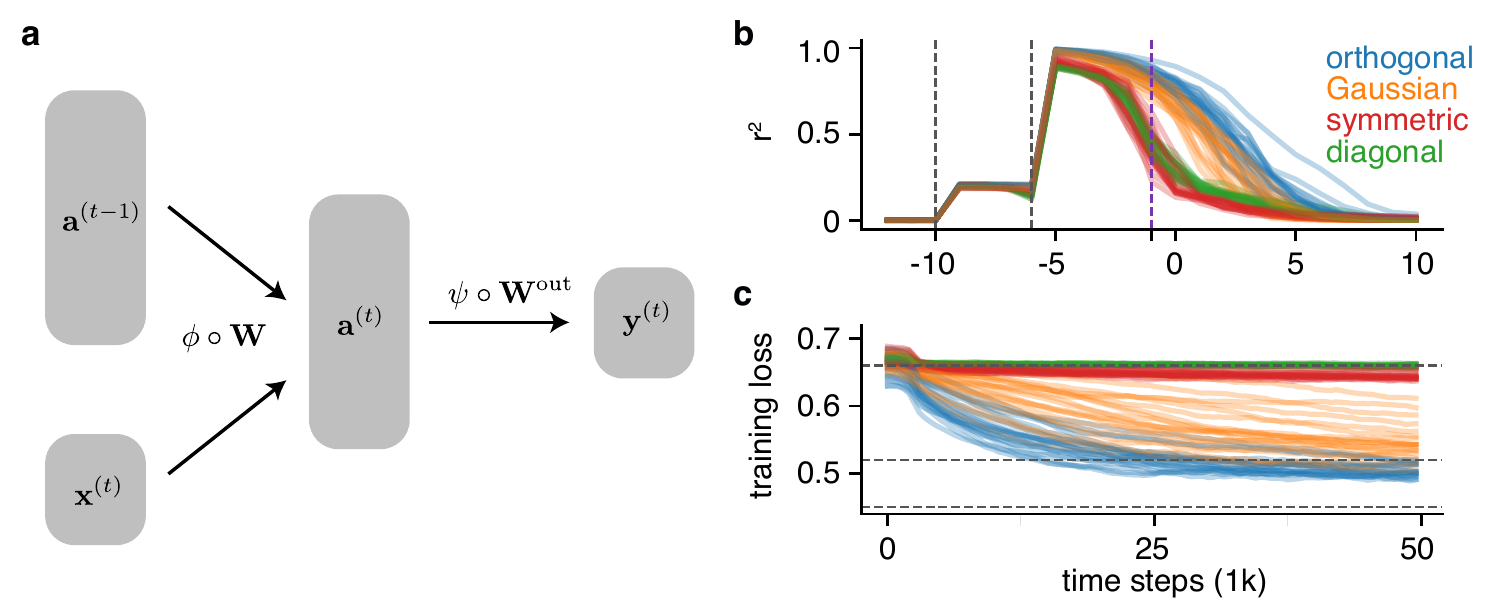}
    \caption{RFLO as a static multi-layer regression. {\bf a)} One time step of recurrent dynamics combine with the output weights to form a single-hidden-layer feedforward network that is trained to predict ${\bf y}^{*(t)}$ from ${\bf \hat{a}}^{(t-1)}$ by the RFLO learning rule ($\psi$ is softmax). {\bf b)} Coefficient of determination $r^2$ between ${\bf a}^{(t + \Delta t)}$ and ${\bf y}^{*(t)}$ as a function of the time shift $\Delta t$ for different methods of generating ${\bf W}$. Each trace is a different random initialization. The cyan dashed line shows the memory at $\Delta t = -1$, corresponding to the input ${\bf a}^{(t-1)}$ of the network in (b). The black dashed lines indicate the lags of the input-output dependencies explicitly included in the task, corresponding to sudden jumps in information about ${\bf y}^{*(t)}$ contained in the hidden state. {\bf c)} Cross-entropy loss over learning by RFLO. The dashed lines indicate benchmarks for learning output statistics, the first dependency, and the second dependency, respectively (see \citealp{pitis2016recurrent} for details). }
    \label{fig:rflo_results}
\end{figure}

We hypothesize that, rather than learning dynamics that actively retain useful bits of the past like RTRL and BPTT, RFLO works by training what is essentially a high-capacity feedforward model to predict labels from the {\bf natural} memory traces of previous inputs contained in the hidden state. This is reminiscient of reservoir computing (\citealp{lukovsevivcius2009reservoir}). We illustrate this idea in the special case of a perfectly discrete network ($\alpha = 1$), where the learning rule still performs remarkably well despite $B^{(t)}_{ij} = \phi'(h^{(t)}_i) \hat{a}^{(t-1)}_j$ containing no network history. As Fig.~\ref{fig:rflo_results}a depicts, ${\bf a}^{(t-1)}$ ultimately maps to ${\bf y}^{(t)}$ via a single-hidden-layer feedforward network parameterized by ${\bf W}$ and ${\bf W}^{\text{out}}$.
The RFLO learning rule in the discrete case corresponds exactly to training ${\bf W}$ by backpropagation:
\begin{equation}
\fracpartial{L^{(t)}}{W_{ij}} = \fracpartial{L^{(t)}}{a^{(t)}_{i}}\fracpartial{a^{(t)}_i}{W_{ij}} = \overline{c}^{(t)}_i \phi'(h^{(t)}_i) \hat{a}^{(t-1)}_j = \overline{c}^{(t)}_i B^{(t)}_{ij}.
\end{equation}
While every learning algorithm additionally trains ${\bf W}^{\text{out}}$ online to best map ${\bf a}^{(t)}$ to ${\bf y}^{*(t)}$, this purely linear model cannot perfectly capture the complex ways that information about past inputs ${\bf x}^{(t')}$, $t' \leq t$ implicit in ${\bf a}^{(t)}$ relates to labels ${\bf y}^{*(t)}$. Adding a hidden layer improves the ability of the network to predict ${\bf y}^{*(t)}$ from whatever evidence of ${\bf x}^{(t')}$ is naturally retained in ${\bf a}^{(t-1)}$, analogous to how a single-hidden-layer feedforward network outperforms a simple softmax regression on MNIST (\citealp{deng2012mnist}).

To empirically validate our explanation, we first show that the strength of natural memory traces in the RNN depends on its recurrent weights. We measure this ``memory" by running an untrained RNN with fixed weights forwards for $20$k time steps of the Add task and calculating the $r^2$ value of a linear regression of ${\bf a}^{(t + \Delta t)}$ onto ${\bf y}^{*(t)}$ for different values of the time shift $\Delta t$. The sudden jumps in information occur as $\Delta t$ passes the time lags of the input-output dependencies explicitly included in the task ($t_1 = 6, t_2 = 10$), followed by a slow decay as the information relevant for predicting ${\bf y}^{*(t)}$ gets corrupted by running the network forwards. The speed of this decay differs by the choice of ${\bf W}$. As Fig.~\ref{fig:rflo_results}b shows, orthogonal and Gaussian ${\bf W}$ seem to best preserve information over time, while symmetric and diagonal ${\bf W}$ lose information quite rapidly, likely due to their real spectra. In separate simulations (Fig.~\ref{fig:rflo_results}c), we trained networks initialized in these ways using the RFLO rule, and only the networks initialized with orthogonal or Gaussian ${\bf W}$ are able to learn at all with RFLO. This validates our hypothesis, that RFLO works by a static prediction of ${\bf y}^{*(t)}$ based on evidence in ${\bf a}^{(t-1)}$, because in cases where this evidence is absent (or at least weak) due to rapid decay, RFLO fails.

\section{Discussion}
\label{discussion}

We have presented a framework for conceptually unifying many recent advances in efficient online training of recurrent neural networks. These advances come from multiple perspectives: pure machine learning, biologically plausible machine learning, and computational neuroscience. We started by articulating a crucial distinction, (a) past facing vs.\ future facing, that divides algorithms by what exactly they calculate. We then presented a few other properties that characterize each new technique: (b) the tensor structure of the algorithm, (c) whether its update requires explicit randomness, and (d) whether its update can be derived in closed form from the relations \eqref{pf_relation}--\eqref{ff_relation} or must be approximated with SGD or some other numerical approach. Along the way, we clarified the relationship between some of the modern online approximations and exact methods. Further, we showed it's possible to create new algorithms by identifying unexplored combinations of properties (a)-(d).

We empirically validated these ideas with synthetic tasks on which all algorithms could perform reasonably well in an online setting, despite using vanilla RNN architecture; gradient descent optimization with fixed learning rate; and no standard machine learning techniques such as gradient clipping, batch/layer normalization, $L^2$ regularization, etc. We saw that training errors roughly cluster according to their categorical distinctions, across tasks and hyperparameter choices. But performance of these clusters differed across tasks: for example, UORO and R-KF-RTRL performed relatively well (poorly) on Add (Mimic), while for DNI and DNI(b) this effect was flipped. We speculate that this can be explained in terms of the task demands: the Add task has a long time horizon (10 time steps) of explicit dependencies for small inputs ($n_{\text{in}} = 1$), while the Mimic task requires more information to be memorized ($n_{\text{in}} = 32$) that decays with the target RNN's forward pass (harder to measure, but likely shorter than 10 time steps). Thus perhaps UORO and R-KF-RTRL produce gradients with a limited amount of information that survives many updates, while DNI and DNI(b) have a larger information capacity but a limited time horizon. Future work should more systematically explore how these different task features interact with the different algorithmic approximations.

Following common practice, we used the pairwise vector alignments of the (approximate) gradients calculated by each algorithm as a way to analyze the precision of different approximations. This similarity turned out to reflect the natural clustering of the algorithms along the axes proposed here. In particular, past-facing approximations had stronger alignment with the exact past-facing gradients calculated by RTRL compared to the exact (up to truncation) future-facing gradients calculated by F-BPTT, and vice versa for the future-facing approximation, DNI. Reassuringly, KeRNL and RFLO, which have the same tensor structure, featured strong alignment.

Importantly, the angular alignment of the gradients did \emph{not} account for task performance: UORO and R-KF-RTRL performed quite well despite their weak alignment with RTRL and BPTT, while KeRNL performed relatively poorly despite its strong alignment with RTRL and BPTT. Analyzing the magnitudes of the gradients partially explained this observation, as UORO and R-KF-RTRL aligned with RTRL more strongly when their gradients were larger in norm. However, this effect was subtle, so probably not enough to account for the performance differences. This exposes a limitation of current ways of analyzing this class of algorithms. There clearly is a need for better similarity metrics that go beyond time-point-wise gradient alignment and instead compare long-term learning trajectories.

\begin{figure}[ht]
    \centering
    \includegraphics{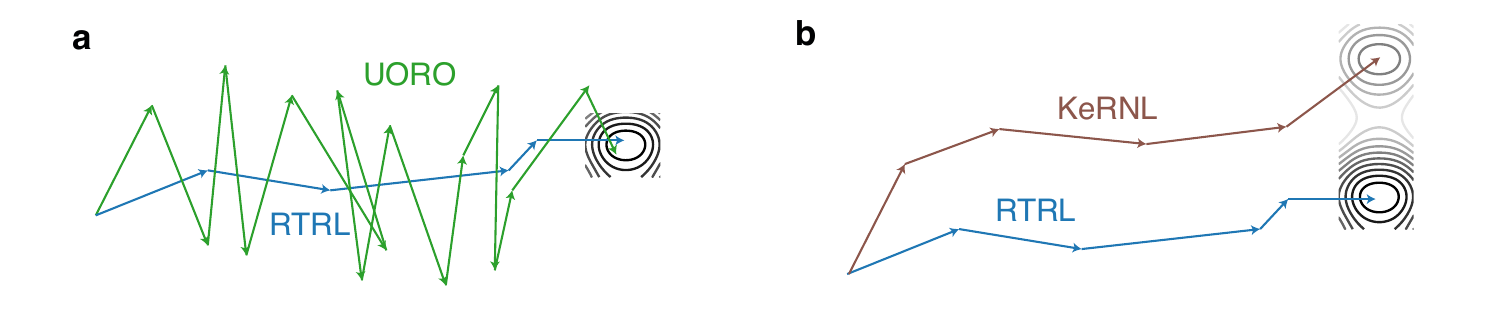}
    \caption{Cartoon illustrating how alignment with RTRL and performance might dissociate. {\bf a)} UORO's noisy estimates of the true gradient are almost orthogonal with RTRL at each time point, but the errors average out over time and allow UORO to find the same solution. {\bf b)} KeRNL aligns more strongly with RTRL at each time point, but errors do not average out, so KeRNL converges to a worse solution.} 
    \label{fig:stochastic_alignment}
\end{figure}

Notably, it is the \emph{stochastic} algorithms that have particularly weak alignment with RTRL---even KF-RTRL, yet its performance is indistinguishable from the exact algorithms. Averaged over many time steps of learning, corresponding to many samples of $\boldsymbol{\nu}$, these stochastic methods do contain complete information about $M^{(t)}_{kij}$ (i.e. are unbiased), but at any one time point the alignment is heavily corrupted by the explicit randomness. In contrast, deterministic approximations, such as KeRNL, RFLO and DNI, may partially align with exact methods by construction, but their errors have no reason to average out, hence their inability to find the same minima as exact methods (Fig.~\ref{fig:stochastic_alignment}). This may also explain why stochastic approximations do not align with each other despite their conceptual similarity.

We chose the simplest possible setup for comparison, using basic RNN architecture and gradient descent, aiming to be as inclusive as possible of various algorithms. Of course, there are alternative online methods that do not naturally fall into this framework. Any online algorithm that exploits a particular RNN architecture necessarily cannot fit into our classification. For example, \cite{ororbia2017learning, ororbia2018online} propose a specialized neural architecture (Temporal Neural Coding Network), whose learning algorithm (Discrepancy Reduction) is naturally local, due to the network structure. In contrast, the algorithms reviewed here were explicitly derived as approximations to nonlocal algorithms (RTRL and BPTT), and their locality ends up being more of a bug than a feature, as evidenced by their impaired performance. From a machine learning perspective, there are limits to general-purpose approximations of this kind, and future progress in online methods will likely come about by further exploring specialized algorithm-architecture pairs.

The same is true from a biological perspective. In the brain, cortical architecture and synaptic plasticity rules evolved together, while physical constraints dictate that plasticity is necessarily local. Still, the details of how local plasticity rules interact with neural circuits remain mysterious, and this is a current focus of research (\citealp{guerguiev2017towards, sacramento2018dendritic, LILLICRAP201982}). Exploring which architectures allow locality to manifest as a consequence, rather than a constraint, of learning is a potentially fruitful point of interaction between artificial intelligence and computational neuroscience.

\acks{We thank Guangyu Yang and James Murray for helpful conversations.}


\bibliography{ref}

\newpage

\appendix
\section*{Appendix A.}\label{sec.lemma}
\subsection*{Lemma for generating rank-1 unbiased estimates}
For completeness, we state the Lemma from \cite{tallec2017unbiased} in components notation. Given a decomposition of a matrix ${\bf M} \in \mathbb{R}^{n \times m}$ into $r$ rank-1 components
\begin{equation}
M_{ij} = \sum_{k=1}^r A_{ik} B_{kj},
\label{lemma:decomposition}
\end{equation}
a vector of i.i.d. random variables $\boldsymbol{\nu} \in \mathbb{R}^r$ with $\mathbb{E}[\nu_k] = 1$,  $\mathbb{E}[\nu_k \nu_{k'}] = \delta_{kk'}$, and a list of $r$ positive constants $\rho_k > 0$, then
\begin{equation}
\tilde{M}_{ij} = \left(\sum_{k=1}^r \rho_k \nu_k A_{ik}\right)\left(\sum_{k=1}^r \rho_k^{-1} \nu_k B_{kj}\right) 
\end{equation}
is a rank-1, unbiased estimate of $M_{ij}$ over the choice of $\boldsymbol{\nu}$.

\section*{Appendix B.}
\subsection*{Implementation details}\label{appendix:hyperparameters}
For reproducibility, we describe in fuller detail our implementation configurations for each simulation. Table~\ref{hyperparameters} shows hyperparameter/configuration choices that apply across all algorithms. Table~\ref{alg_hyperparams} shows the algorithm-specific hyperparameter choices we made for each task. In Table~\ref{hyperparameters}, we reference sub-matrices of ${\bf W} = [{\bf w}^{\text{rec}}, {\bf w}^{\text{in}}, {\bf b}^{\text{rec}}]$ and ${\bf W}^{\text{out}} = [{\bf w}^{\text{out}}, {\bf b}^{\text{rec}}]$, since they are initialized differently.

\begin{table}[hb]
\begin{tabular}{|l|l|l|}
\hline
\textbf{hyperparameter}                     & \textbf{value}                                 & \textbf{explanation}                                                                 \\ \hline
learning rate                               & $10^{-4}$                                      & learning rate for SGD w.r.t. ${\bf W}$ and ${\bf W}^{\text{out}}$         \\ \hline
$n$                                         & 32                                             & number of hidden units in the network                                                \\ \hline
$\phi$                                      & tanh                                           & nonlinearity used in RNN forward dynamics                                            \\ \hline
init. ${\bf w}^{\text{in}}$                 & $\sim \mathcal{N}(0, 1/\sqrt{n_{\text{in}}})$  & initial value for input weights                                                      \\ \hline
init. ${\bf w}^{\text{rec}}$                & rand. orth.                                    & initial value for recurrent weights \\ \hline
init. ${\bf b}^{\text{rec}}$                & 0                                              & initial value for recurrent bias                                                   \\ \hline
init. ${\bf w}^{\text{out}}$                & $\sim \mathcal{N}(0, 1/\sqrt{n})$              & initial value for output weights                                                     \\ \hline
init. ${\bf b}^{\text{out}}$                & 0                                              & initial value for output bias                                                      \\ \hline
init. ${\bf W}^{\text{FB}}$                 & $\sim \mathcal{N}(0, 1/\sqrt{n_{\text{out}}})$ & value for fixed feedback weights used in DNI(b)                              \\ \hline
init. ${\bf b}^{\text{rec}}_{\text{targ.}}$ & $\sim \mathcal{N}(0, 0.1)$                     & initial value for target recurrent bias in Mimic                               \\ \hline
init. ${\bf b}^{\text{out}}_{\text{targ.}}$ & $\sim \mathcal{N}(0, 0.1)$                     & initial value for target output bias in Mimic                                  \\ \hline
\end{tabular}
\caption{Default hyperparameter choices for the RNN independent of learning algorithm.}
\label{hyperparameters}
\end{table}

\begin{table}[ht]
\begin{tabular}{|l|l|l|l|l|l|}
\hline
\textbf{algorithm} & \textbf{initial values}                                                                      & \textbf{$\boldsymbol{\nu}$ dist.} & \textbf{LR} & \textbf{pert. $\sigma$} & \textbf{$T$} \\ \hline
UORO               & $A^{(0)}_k \sim \mathcal{N}(0, 1), B^{(0)}_{ij} \sim \mathcal{N}(0, 1)$                      & unif. $\{-1, 1\}$                 &             &                         &     \\ \hline
KF-RTRL            & $A^{(0)}_j \sim \mathcal{N}(0, 1), B^{(0)}_{ki} \sim\mathcal{N}(0, 1/\sqrt{n})$              & unif. $\{-1, 1\}$                 &             &                         &     \\ \hline
R-KF-RTRL          & $A^{(0)}_i \sim \mathcal{N}(0, 1), B^{(0)}_{kj} \sim\mathcal{N}(0, 1/\sqrt{n})$              & unif. $\{-1, 1\}$                 &             &                         &     \\ \hline
KeRNL              & $A^{(0)}_{ki} = \delta_{ki}, B^{(0)}_{ij} = 0, \alpha^{(0)}_i = 0.8$                         &                                   & $5$         & $10^{-3}$               &     \\ \hline
DNI                & $A^{(0)}_{li} \sim \mathcal{N}(0, 1/\sqrt{n})$                                               &                                   & $10^{-3}$   &                         &     \\ \hline
DNI(b)             & $A^{(0)}_{li} \sim \mathcal{N}(0, 1/\sqrt{n}), \mathcal{J}^{(0)}_{ij} = W^{\text{rec}}_{ij}$ &                                   & $10^{-3}$   &                         &     \\ \hline
F-BPTT             &                                                                                              &                                   &             &                         & 10  \\ \hline
\end{tabular}
\caption{Hyperparameter choices specific to individual algorithms.}
\label{alg_hyperparams}
\end{table}

Some miscellaneous implementation details below:
\begin{itemize}
\itemsep0em
    \item For the Add task in the $\alpha = 1$ condition, we changed the DNI/DNI(b) learning rate to $5 \times 10^{-2}$ for $A_{li}$ and $10^{-2}$ for $\mathcal{J}_{ij}$ (DNI(b)). In other cases, the learning rates for $A_{li}$ and $\mathcal{J}_{ij}$ are identical.
    \item There are two appearances of the synthetic gradient weights $A_{li}$ in Eq.~\eqref{sg_update}. Although we wrote them as one matrix ${\bf A}$ for brevity, in implementation we actually keep two separate values, ${\bf A}$ and ${\bf A}^*$, the latter of which we use for for the right-hand appearance $A_{l'm}$ (specifically to calculate the bootstrapped estimate of the SG training label). We update ${\bf A}$ every time step but keep ${\bf A}^*$ constant, replacing it with the latest value of ${\bf A}$ only once per $\tau \in \mathbb{N}$ time steps. This integer $\tau$ introduces another hyperparameter, which we choose to be 5. (Inspired by an analogous technique used in deep Q-learning from \citealp{mnih2015human}.)
    \item In the original paper, \cite{roth2018kernel} use $(1 - \exp(-\gamma_i))$ rather than $\alpha_i$ as a temporal filter for $B^{(t)}_{ij}$. We made this change so that $\alpha_i$ makes sense in terms of the $\alpha$ in the forward dynamics of the network and RFLO. Of course, these are equivalent via $\gamma_i = -\log(1 - \alpha_i)$, but the gradient w.r.t. $\alpha_i$ must be rescaled by a factor of $1/(1 - \alpha_i)$ to compensate.
    \item For KeRNL, there is a choice for how to update the eligibility trace (Eq.~\ref{e_trace}): one can scale the right-hand term $\phi'(h^{(t)}_i)\hat{a}^{(t-1)}_j$ by either the learned timescale $\alpha_i$ or the RNN timescale $\alpha$. We chose the latter because it has stronger empirical performance and it theoretically recovers the RTRL equation under the approximating assumptions about ${\bf A}$.
    \item Perturbations for calculating gradients for $A_{ki}$ and $\alpha_i$ in KeRNL are sampled i.i.d. $\zeta_i \sim \mathcal{N}(0, \sigma)$.
    \item In our implementation of the Add task, we use $n_{\text{in}} = n_{\text{out}} = 2$ for a ``one-hot" representation of the input $x^{(t)} \in \{0, 1\}$ and label $y^{*(t)} \in \{0.25, 0.5, 0.75, 1\}$, such that ${\bf x}^{(t)} = [x^{(t)}, 1 - x^{(t)}]$ and ${\bf y}^{*(t)} = [y^{(t)}, 1 - y^{(t)}]$.
    \item In our implementation of Mimic, the target RNN was initialized in the same way as the RNNs we train, with the exception of the recurrent and output biases (see Table~\ref{hyperparameters}).
\end{itemize}
\vskip 0.2in

\end{document}